%% file: emnlp2023.tex
\pdfoutput=1

\documentclass[11pt]{article}

\usepackage[]{EMNLP2023}

\usepackage{times}
\usepackage{latexsym}

\usepackage{comment}
\usepackage{rotating}
\usepackage[utf8]{inputenc}
\usepackage{times}
\usepackage{latexsym}
\usepackage{rotating,tabularx,siunitx}

\usepackage{float}
\usepackage{graphicx}
\usepackage{subfigure}
\usepackage{caption}
\usepackage{amsmath,amssymb}
\usepackage{mathtools}
\usepackage{booktabs}
\usepackage{multirow}
\usepackage{multicol}
\usepackage{siunitx}
\usepackage{adjustbox}
\usepackage{pifont}

\usepackage{enumitem}

\newcommand{\RNum}[1]{\uppercase\expandafter{\romannumeral #1\relax}}

\usepackage[T1]{fontenc}

\usepackage[utf8]{inputenc}

\usepackage{microtype}

\usepackage{inconsolata}
\newcommand{\ust}{\ensuremath{^\spadesuit}}
\newcommand{\nvidia}{\ensuremath{^\dagger}}

\title{Persona Knowledge-Aligned Prompt Tuning Method for Online Debate}

\author{Chunkit Chan\ust \ \ \ \ Cheng Jiayang\ust \ \ \ \ Xin Liu\ust \ \ \ \ Yauwai Yim\ust \\
\ \ \ \ \textbf{Yuxin Jiang}\ust \ \ \ \ \textbf{Zheye Deng}\ust \ \ \ \ \textbf{Haoran Li\ust} \\ 
\ \ \ \ \textbf{Yangqiu Song\ust} \ \ \ \ \textbf{Ginny Y. Wong\nvidia} \ \ \ \ \textbf{Simon See\nvidia}\\
\ust The Hong Kong University of Science and Technology \\
\nvidia NVIDIA AI Technology Center (NVAITC)\\
\texttt{\{ckchancc, yqsong\}@cse.ust.hk} \ \ \ \ 
}

\begin{document}
\maketitle
\input{sections/abstract.tex}
\input{sections/introduction.tex}
\input{sections/persona.tex}

\input{sections/model.tex}

\input{sections/experiments.tex}

\input{sections/related_work.tex}
\input{sections/conclusion.tex}

\input{sections/limitation.tex}

\input{sections/ethics_statement.tex}
\section*{Acknowledgements}
The authors of this paper were supported by the NSFC Fund (U20B2053) from the NSFC of China, the RIF (R6020-19 and R6021-20) and the GRF (16211520 and 16205322) from RGC of Hong Kong. We also thank the support from NVIDIA AI Technology Center (NVAITC).

\bibliography{anthology,custom}
\bibliographystyle{acl_natbib}

\clearpage
\appendix

\input{sections/appendix.tex}

\end{document}

%% file: sections/abstract.tex
\begin{abstract}
Debate is the process of exchanging viewpoints or convincing others on a particular issue. Recent research has provided empirical evidence that the persuasiveness of an argument is determined not only by language usage but also by communicator characteristics. Researchers have paid much attention to aspects of languages, such as linguistic features and discourse structures, but combining argument persuasiveness and impact with the social personae of the audience has not been explored due to the difficulty and complexity. We have observed the impressive simulation and personification capability of ChatGPT, indicating a giant pre-trained language model may function as an individual to provide personae and exert unique influences based on diverse background knowledge. Therefore, we propose a persona knowledge-aligned framework for argument quality assessment tasks from the audience side. This is the first work that leverages the emergence of ChatGPT and injects such audience personae knowledge into smaller language models via prompt tuning. The performance of our pipeline demonstrates significant and consistent improvement compared to competitive architectures.
\end{abstract}

%% file: sections/introduction.tex
\section{Introduction}
In the field of Natural Language Processing (NLP) and Computational Argumentation, there is a burgeoning research interest in studies to develop computational methods that can automatically assess the qualitative characteristics of arguments.
The impact and persuasiveness of the argument are crucial and pivotal qualitative characteristics, and substantial research has been conducted to develop computing methodologies for identifying the impact and the persuasiveness of a natural language argument in public debate forums~\cite{DBLP:conf/emnlp/HabernalG16,DBLP:conf/acl/HabernalG16,DBLP:conf/www/TanNDL16,DBLP:conf/naacl/DurmusC18,DBLP:journals/tacl/SimpsonG18,DBLP:conf/emnlp/LiDC20}.
Nevertheless, estimating the impact or persuasiveness of an argument covering various debate topics requires more extensive knowledge
than merely comprehending the surface semantic meaning of an argument in online debate forums. In argumentation mining, \citet{DBLP:journals/tacl/LauscherWGG22} define the term \textbf{knowledge} as \textit{any kind of normative information that is considered to be relevant for solving a task at hand and that is not given as task input itself}.

\begin{figure}[t]
    \centering
    \includegraphics[width=\linewidth]{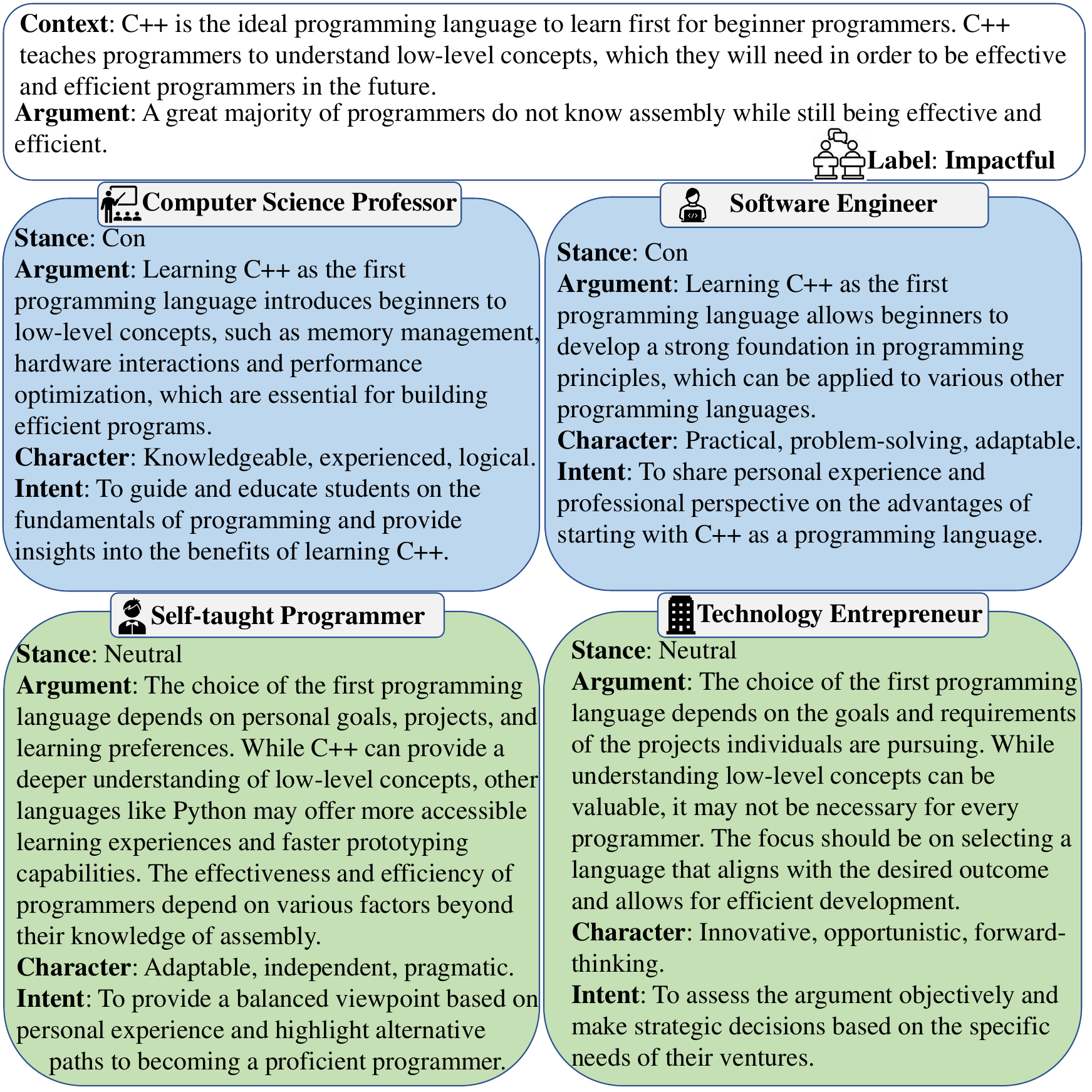}
    \vspace{-0.7 cm}
    \caption{A data example from \textit{Kialo} and diverse audience personas generated by ChatGPT on this online debate topic. The \textit{Context} indicates the previous historical arguments from other users. The \textit{Argument} indicates the current argument or statement from the users.}
    \label{fig:DataInstance_Introduction}
    \vspace{-0.7 cm}
\end{figure}

Traditional works in argument assessment tasks have studied various aspects of knowledge~\cite{DBLP:journals/tacl/LauscherWGG22}. Among them, the impact and persuasiveness of arguments are inextricably linked not only to the linguistic attributes of the language~\cite{DBLP:conf/acl/WeiLL16} but also to the traits of the communicators, including the source (speakers)~\cite{DBLP:conf/www/DurmusC19}, 
the prior beliefs~\cite{DBLP:conf/naacl/DurmusC18},
argument structure~\cite{DBLP:conf/emnlp/LiDC20}, and the influence of discourse contexts~\cite{DBLP:conf/acl/LiuOSJ20}.
However, previous works have not well explored the analysis of the social personae of the audience in a computational manner, except by annotating human subjects on the speaker side~\cite{DBLP:conf/eacl/WalkerALW17}.
The recent computational studies for personae in Large Language Models (LLM) underscore the significance of personality information~\cite{jiang2023personallm,li2024personal}.
Furthermore, research in social psychology has identified the factors of argument persuasiveness, one of which is the \textbf{audience}~\cite{petty1986elaboration,johnson2013role, DBLP:journals/patterns/AlshomaryW21}, as a substantial amount of content is not expressed explicitly but resides in the mind of the audience~\cite{DBLP:journals/argcom/Moens18}, and the impact and persuasiveness of arguments are highly dependent on the audience.

Figure~\ref{fig:DataInstance_Introduction} illustrates various audience personae having various stances and interpretations on the same stated context and argument.
An individual's persona exerts significant influence on his/her own background knowledge (e.g., prior beliefs) and personality (e.g., roles' characters and intents)~\cite{DBLP:conf/naacl/DurmusC18}. Hence, diverse audience personas formulate different stances and arguments on a particular debate argument according to their characters and intentions.
Therefore, we adopt this focus since persona knowledge on the audience side plays a crucial role in forming stances and viewpoints about controversial topics and ultimately helps to determine the impact and persuasiveness of the argument.

Nevertheless, the challenge lies in the high level of difficulty and complexity associated with acquiring the audience's persona knowledge, and the manual collection imposes difficulty on the scalability of persona knowledge for various argument assessment tasks. 
Numerous works have successfully elicited knowledge from large language models instead of retrieving it on the knowledge graph~\cite{
DBLP:conf/emnlp/WangHQSWLG22}.
ChatGPT~\cite{openai2022chatgpt} has demonstrated the ability to act in diverse roles given the instructions and applied for different tasks and areas~\cite{DBLP:journals/corr/abs-2305-14325,
DBLP:journals/corr/abs-2302-04023}, especially simulating an open world and the roles~\cite{DBLP:journals/corr/abs-2304-03442}.
Hence, in this paper, we utilize ChatGPT imitation of various audience roles on each debate topic and argument, prompting persona knowledge through the tailored prompt to explore its influence on the argument assessment task. Furthermore, we proposed a persona knowledge-aligned framework for aligning the persona knowledge from LLM (i.e., ChatGPT) to a smaller language model (i.e., FLAN-T5) via prompt tuning to undertake the argument assessment task. The persona knowledge infuses into the tunable prefix prompt tokens without altering the pre-trained model representations.
Our contributions are summarized as follows:
{\begin{itemize}[leftmargin=*]
    \item To the best of our knowledge, this is the first work that explores and aligns audience persona knowledge into pre-trained language models via prompt tuning on the argument quality assessment task\footnote{The source code is available at~\url{https://github.com/HKUST-KnowComp/PersonaPrompt}}.
    \item We designed a framework to elicit human-validated audience persona knowledge from a large language model (i.e., ChatGPT) to help determine the impact and persuasiveness of the argument. 
    \item We conduct extensive experiments and thorough ablation studies to discuss the necessity and effectiveness of the various tailored dimensions of persona knowledge and the proposed method.
\end{itemize}}

%% file: sections/persona.tex
\section{Persona Knowledge}
\subsection{Persona Knowledge Generation}
\paragraph{Data Argument and Context} 
There are two forms of debate presented online. One is the arguments are typically structured as a series of rounds, with each round featuring an utterance from the PRO side and one from the CON side\footnote{To maintain consistency in methodology, we call a round with arguments from two debaters as one argument.} (e.g., \textit{DDO} dataset~\cite{DBLP:conf/acl/DurmusC19, DBLP:conf/emnlp/LiDC20}). 
In contrast, open debate platforms like \textit{Kialo}\footnote{\url{https://www.kialo.com/}} often adopt a more informal approach, allowing individuals to express their stances and argument claims to provide support or opposition to various topics or arguments, where the process can be organized as a debate tree.
We define the \textbf{argument claim} denoted as $\mathcal{A}$ to be the argumentative and persuasive text to express an idea for the audience and regard other relevant arguments in previous rounds or from other speakers as the \textbf{context} $\mathcal{C}$, $\mathcal{C}$ = $(\mathcal{C}^0, \mathcal{C}^1, \cdots, \mathcal{C}^{l})$ where $l$ is context length and $\mathcal{C}^{l}$ is the parent argument of $\mathcal{A}$.

\begin{figure}[!t]
    \vspace{-0.2cm}
    \centering
    \includegraphics[width=1.05\linewidth]{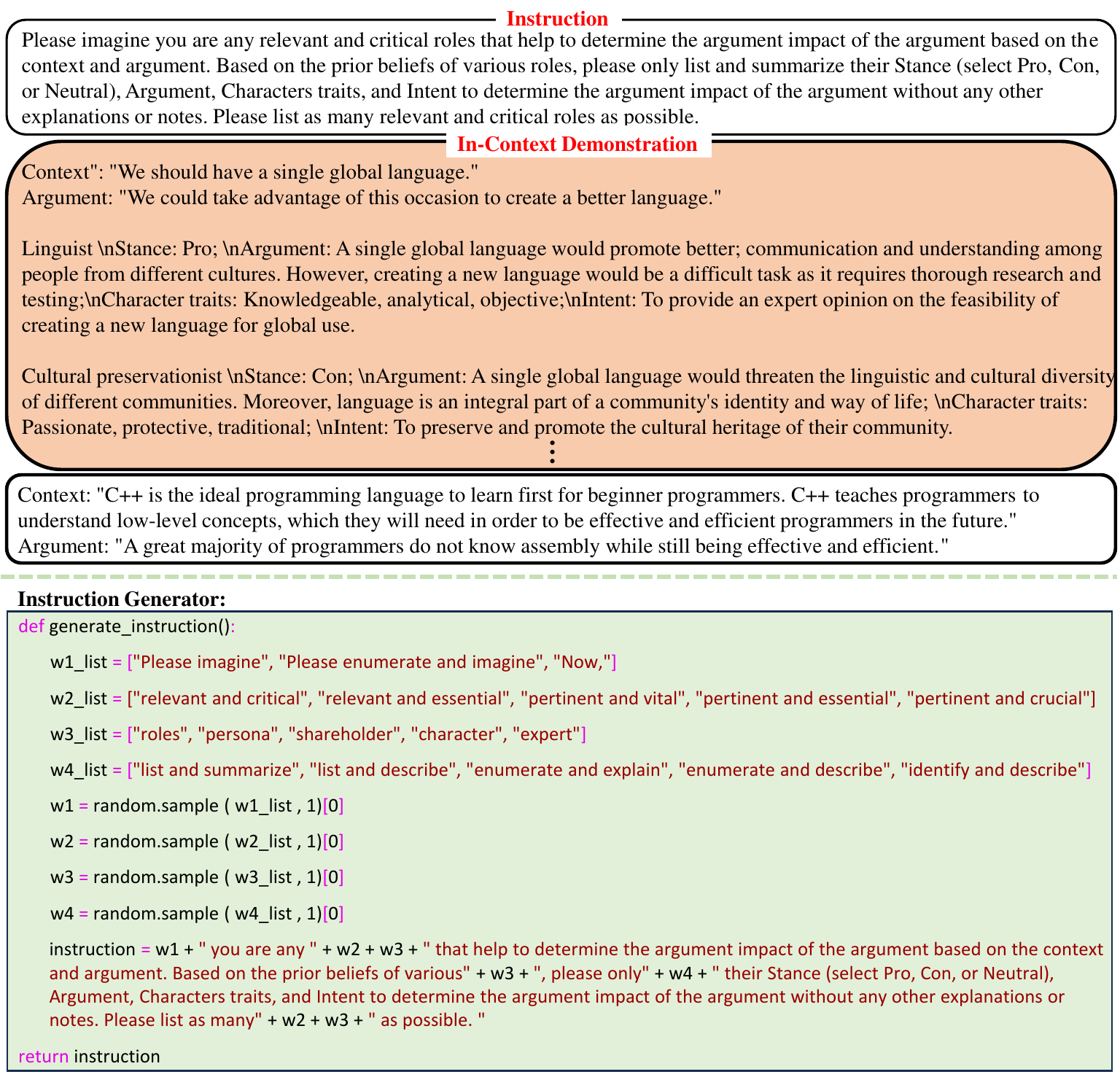}
    \vspace{-0.7cm}
    \caption{The upper portion is a prompt template for eliciting the persona knowledge from ChatGPT, and the bottom portion is the randomized instruction generator.}
    \label{fig:ChatGPT_PromptTemplate}
    \vspace{-0.5cm}
\end{figure}

\paragraph{Dimensions of Persona Knowledge}
We notice that \citet{moore2019persona} proposed five common dimensions of persona: Public, Mediatized, Performative, Collective, and VARP (\textit{Values, Agency, Reputation, Prestige}) dimensions.
However, those dimensions are too general for debate arguments, which may not be specific and adaptable to broader debate topics. To construct efficient and task-specific representations of persona knowledge on the argument quality assessment task, we intuitively and meticulously design four potential dimensions for each persona knowledge instance as follows:

\noindent \textbf{Persona Stance} This dimension describes the stance of an audience persona (i.e., \textit{Con}, \textit{Pro}, or \textit{Neutral}) regarding the given argument and context. 

\noindent \textbf{Persona Argument} This dimension presents the audience persona argument that supports their stance, according to their own characters and intentions.

\noindent \textbf{Persona Characters} This dimension describes intrinsic character traits that a persona is likely to exhibit. 

\noindent \textbf{Persona Intent} This dimension outlines the external action or outcome that an audience persona intends to achieve or accomplish in the forthcoming period. Given that diverse audience personas take different stances and arguments on a specific debate argument according to their characters and intentions, this dimension is an integral part of persona knowledge in this work.

\paragraph{Persona Knowledge Generation}
To ensure the high quality and diversity of elicited multi-dimensional personae from ChatGPT\footnote{Disclaimer: All generated persona knowledge reflects the selection and reporting biases~\cite{gordon2013reporting} of ChatGPT, which could sometimes be stereotypical and do not represent the views of the authors.} and mitigate the issues of LLMs sensitive to instruction and few-shot examples~\cite{DBLP:conf/nips/PerezKC21}, we have customized the dynamic prompting template and introduced randomization in the prompts. This is achieved by \textbf{(I)} manually creating a collection of semantically similar instructions and randomly sampling from the instruction set each time, \textbf{(II)} creating an initial in-context examples pool, and dynamically sampling in-context examples for each input.
The initial in-context examples pool includes 100 manually refined persona knowledge for 10 well-chosen instances, which are crafted to cover as many relevant, critical, and diverse personas as possible.
The prompt template is displayed in Figure~\ref{fig:ChatGPT_PromptTemplate}, and an example of the persona knowledge generated by ChatGPT is presented in Figure~\ref{fig:DataInstance_Introduction}. For a given context $c_i \in \mathcal{C}$ and argument claim of $a_i \in \mathcal{A}$, by employing large language model $\mathcal{M}$, we sample persona knowledge $p_i \in \mathcal{P}$: 
\begin{align}
    p_i \sim \mathcal{M}(p_i \mid c_i, a_i)
\end{align}
where $i$ indicates $i$-th instance of the dataset $\mathcal{D}=\left\{\left(x_{i},y_{i}\right)\right\}_{i=1}^{|\mathcal{D}|}$ and $x_{i}=\left\{c_i, a_i\right\}$.
It is noteworthy that the persona knowledge generated by ChatGPT and GPT-4 exhibits significant similarity. We have opted for ChatGPT to generate all persona knowledge to optimize cost efficiency, and GPT-4 generated persona knowledge shown in Appendix~\ref{sec:More_case_GPT-4}.

\begin{figure}[!t]
    \centering
    \vspace{-0.2cm}
    \includegraphics[width=0.85\linewidth]{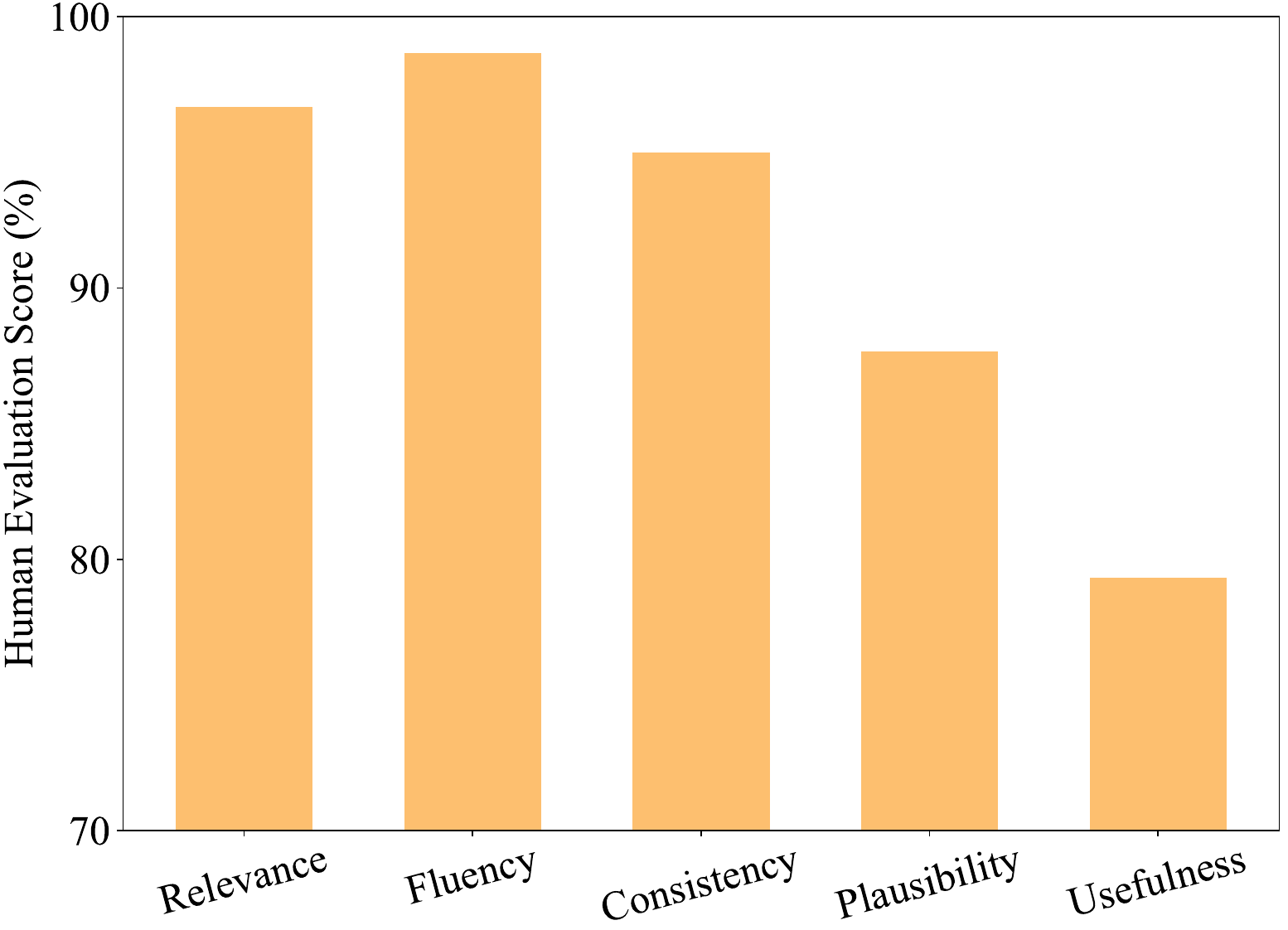}
    \vspace{-0.15in}
    \caption{Human validation of five aspects of persona knowledge elicited from ChatGPT. The human evaluation score for the Harmfulness aspect is zero, indicating that no harmful or toxic language is found in the 1,000 sampled personae, which is omitted from the figure.}
    \label{fig:Human_Evaluation}
    \vspace{-0.5cm}
\end{figure}

\begin{figure*}[!t]
\centering
\includegraphics[width=\textwidth]{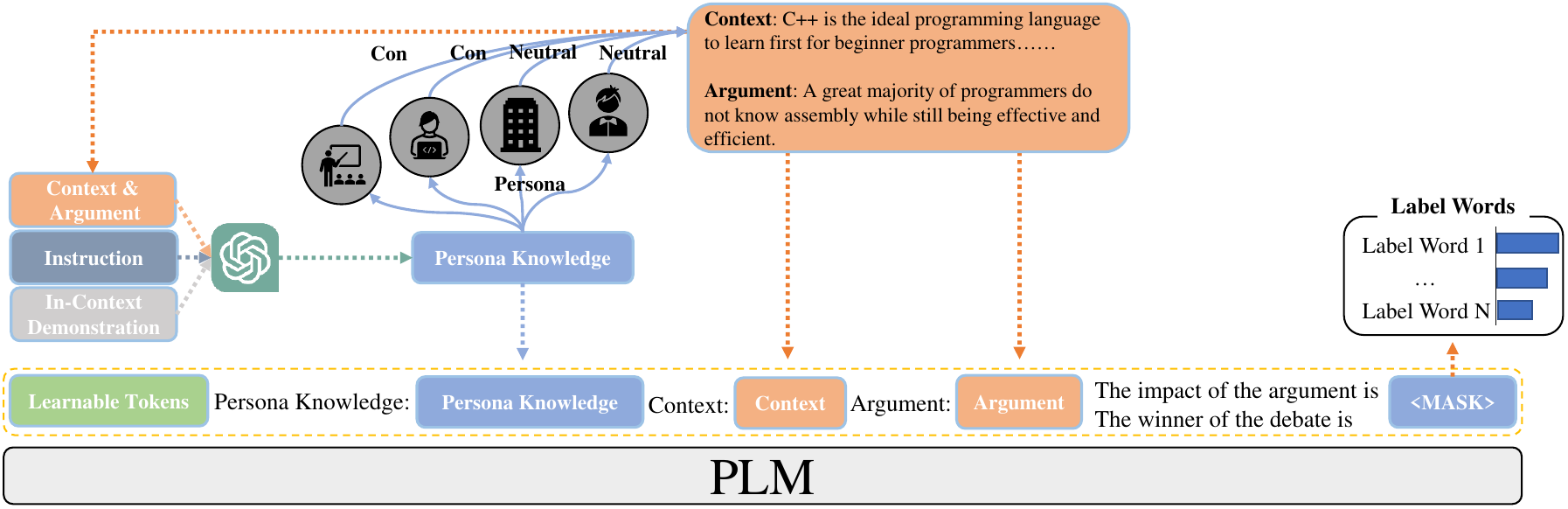}
\vspace{-0.7cm}
\caption{
Overview of the PresonaPrompt framework.
}
\label{fig:DiscoursePrompt_Architecture}
\vspace{-0.5cm}
\end{figure*}

\subsection{Human Validation}
To assess the quality, effectiveness, and helpfulness of generated persona knowledge and to address potential hallucination issues in the generated content~\cite{DBLP:journals/corr/abs-2305-11747}, we conduct a human validation to complement the experimental results. 
We score each persona and the corresponding four dimensions of generated knowledge in the following intrinsic and extrinsic criteria:
(1) \textbf{Relevance} that determines whether the Roles, Argument, and Intent are relevant to the data argument and context;
(2) \textbf{Fluency} that assesses the fluency and understandability of the Roles, Argument, and Intent;
(3) \textbf{Consistency} that evaluates whether the Intent and Argument are consistent with Stance;
(4) \textbf{Plausibility} that gauges the reasonableness and plausibility of the Intent and Argument;
(5) \textbf{Usefulness} that measures whether the generation helps in determining the persuasiveness of the data argument;
(6) \textbf{Harmfulness} that estimates whether the generated knowledge includes harmful and toxic language or words.

We randomly sample 1,000 persona roles from the 250 debates in the testing set, and five annotators are asked to evaluate every role in these dimensions, yielding a total of 30,000 ratings for all sampled knowledge (1,000 persona roles $\times$ 6 aspects $\times$ 5 annotators). We take the majority vote among five votes as the final result for each persona knowledge. The Inter-Annotator Agreement (IAA) score is 73.33\%, computed using pairwise agreement proportion, and Fleiss's Kappa~\cite{fleiss1971measuring} is 0.45. 
The average scores are in Figure~\ref{fig:Human_Evaluation}.
The relevance, fluency, consistency, and harmfulness aspects receive the higher agreement, while the plausibility and usefulness aspects obtain relatively lower agreement among annotators compared with other dimensions but also obtain 87\% and 79\% IAA scores. Notably, the harmfulness aspect in these 1,000 sample persona knowledge is zero, and a zero score means no harmful or toxic language is detected in the generation. It may be attributed to the ChatGPT fine-tuned with reinforcement learning from human feedback (RLHF) approach, which prevents ChatGPT from generating harmful language without deliberate attacks~\cite{DBLP:conf/nips/ChristianoLBMLA17}.

%% file: sections/model.tex
\section{Persona Knowledge Aligned Prompt Framework}
\noindent \textbf{Problem Definition.}
Despite the differences in debate forms, the primary objective of debaters remains to persuade the audience effectively.
Therefore, our aim is to utilize machine learning methods to predict the winner of a debate based on the persuasiveness of their arguments.
This approach allows us to frame argument assessment tasks as classification problems giving an argument claim $\mathcal{A}$ and its corresponding context $\mathcal{C}$,
predict the label $\mathcal{Y} \in$ \{\textit{Con, Pro}\} in the debate form of DDO benchmark, while $\mathcal{Y} \in$ \{\textit{Impactful, Medium Impact, Not Impact}\} in Kialo debate forms. Therefore, this task is to find out the proper winner or plausible impact level based on the persuasiveness:

\scalebox{0.91}{\parbox{1.10\linewidth}{
\begin{align}
    y_{i}^{*}=\arg \max _{y_{i}^{j}} \text{Pr}\left(y_{i}=y_{i}^{j} \mid x_{i}^{j}\right),
\end{align}
}}
where $y_{i}^{*}$ is the most persuasive winner or most reasonable impact level, and $j$ indicates the $j$-th label among all labels.

\subsection{PersonaPrompt}
To predict the label $y_i$ for each instance $x_i = \{(c_i, a_i)\}$, we append the corresponding audience persona knowledge $p_i$ to each instance. Then, we leverage a human-tailored template $\mathcal{T}(\cdot)$ to convert the data instances and the persona knowledge to the prompt input $\tilde{x}_i = \mathcal{T}(p_i, x_i)$ and a verbalizer $\mathcal{V}(\cdot)$ to map a set of label words to class labels. Figure~\ref{fig:DiscoursePrompt_Architecture} illustrates the overall framework.

\noindent \textbf{Knowledge-Aligned Template}
The crafted template includes necessary discrete tokens and learnable continuous tokens.
As shown in Figure~\ref{fig:DiscoursePrompt_Architecture}, we utilize indicators to separate the ``context'' and ``argument'' and instruct the models to predict either the winner of the debate or the potential impact level of an argument.
In addition, we incorporate persona knowledge generated by ChatGPT as the ``background'' preceding the context, which aligns persona knowledge from a large language model to the small model to enhance the comprehension ability 
on the tasks, providing a broader and more comprehensive perspective on the debate process. 
We also appended 20 learnable continuous tokens at the beginning of the input template, allowing them to be updated through backpropagation.

\noindent \textbf{Verbalizer}
A traditional verbalizer $\mathcal{V}(\cdot)$ is a mapping function ($\mathcal{V}:  \mathcal{Y} \rightarrow \mathcal{Z} $) designed for bridging the set of answer token $\mathcal{Z}$ to the class label set $\mathcal{Y}$~\cite{DBLP:journals/corr/abs-2107-13586}. Normally, by using the prompt template and the function $\mathcal{V}(\cdot)$, the probability distribution over $\mathcal{Y}$ can be formalized as the probability distribution over $\mathcal{Z}$ at the masked position, i.e., $\text{Pr}(y_i | \tilde{x}_i) = \text{Pr}(\mathcal{V}(y_i)|\tilde{x}_i) = \text{Pr}(z_i| \tilde{x}_i)$.
To explicitly exhibit the contribution and effectiveness of persona knowledge in our model, we simplify the verbalizer function, which treats the original class label with lowercase as the label words (e.g., "Con" to "con").
Thus, we predict the label by choosing the higher probability answer token:\\
\scalebox{0.91}{\parbox{1.10\linewidth}{
\begin{align}
    y_{i}^{*}=\arg \max _{z_{i}^{j}} \text{Pr}\left(z_{i}=z_{i}^{j} \mid \tilde{x}_i^{j}\right).
\end{align}
}}
The final learning objective of PersonaPrompt is to maximize\\
\scalebox{0.91}{\parbox{1.10\linewidth}{
\begin{align}
    \mathcal{J} = \frac{1}{|\mathcal{D}|} \sum_{(x_i, y_i) \in \mathcal{D}} \log \text{Pr}\left(z_{i}=z_{i}^{j} \mid \tilde{x}_i^{j}\right).
\end{align}
}
}

%% file: sections/experiments.tex
\section{Experimental Setting}
\subsection{Task Datasets}
To validate the effectiveness of persona knowledge in debate tasks, we conducted experiments on two tasks: argument impact classification in the \textit{Kialo} dataset \cite{DBLP:conf/emnlp/DurmusLC19} and argument persuasion prediction in the \textit{DDO} dataset~\cite{DBLP:conf/acl/DurmusC19, DBLP:conf/emnlp/LiDC20}. The Kialo dataset collected various topics and categorized the user votes into three impact classes (\textit{Not Impactful, Medium Impact,} and \textit{Impactful}) based on agreement and the number of valid votes to reduce noise. The dataset statistics are presented in Table~\ref{table:stat_label} in Appendix~\ref{sec:appendix_data_statistics}. The \textit{DDO} dataset is used for the argument persuasion prediction task. The task involves predicting the winner who presented more convincing arguments in a debate. Each debate consists of multiple rounds, with each round featuring an utterance from the PRO side and one from the CON side. More details can be found in Appendix~\ref{sec:appendix_data_statistics}.

\subsection{PresonaPrompt Implementation Details}
We employ Flan-T5~\cite{DBLP:journals/corr/abs-2210-11416}, which uses instructions to tune the T5 model~\cite{DBLP:journals/jmlr/RaffelSRLNMZLL20}, as the PLM backbone in \textbf{PersonaPrompt}. For fair comparison with various baselines, we used Flan-T5-base as the primary model.
In general, the overall configuration follows the setup in \citet{DBLP:conf/emnlp/LesterAC21}, and more configuration details can be found in Appendix~\ref{sec:appendix_discoPrompt_implementation_details}.
Following previous works~\cite{DBLP:conf/emnlp/DurmusLC19, DBLP:conf/acl/LiuOSJ20, DBLP:conf/emnlp/LiDC20}, we report the Macro-F1 score for the argument impact classification task and ablation study, while accuracy is reported for the argument persuasion prediction task. All experiments are conducted using 2 $\times$ NVIDIA V100 (32GB) GPUs.

\subsection{Baselines}
This paper mainly adopts two categories of competitive baselines for the \textit{Kialo} dataset and the \textit{DDO} dataset. The first category consists of the previous state-of-the-art baselines, such as LR~\cite{DBLP:conf/emnlp/DurmusLC19}, SVM~\cite{DBLP:conf/acl/DurmusC19}, BiLSTM~\cite{DBLP:conf/acl/LiuOSJ20}, HAN-BiLSTM~\cite{DBLP:conf/acl/LiuOSJ20}, BERT~\cite{DBLP:conf/emnlp/DurmusLC19}, and DisCOC~\cite{DBLP:conf/acl/LiuOSJ20}.
The other category involves the fine-tuned Flan-T5 models to illustrate the performance gain of prompt tuning. Additionally, we include general Prefix-Tuning~\cite{DBLP:conf/acl/LiL20} as well as Prompt-Tuning~\cite{DBLP:conf/emnlp/LesterAC21}. 
The details of the baselines and their implementation are listed in Appendix~\ref{sec:appendix_experiments}.

\begin{table*}[!t]
    \centering
    \small
    \scalebox{0.75}{\begin{tabular}{l|c c c}
    \toprule
    \multicolumn{1}{c|}{\textbf{Model}} &  \multicolumn{1}{c}{\textbf{Precision}} &  \multicolumn{1}{c}{\textbf{Recall}} &  \multicolumn{1}{c}{\textbf{Macro F1}} \\ 
    \midrule
    \textsc{Majority} & 19.43 & 33.33 & 24.55 \\
    \textsc{SVM}~\cite{DBLP:conf/emnlp/DurmusLC19} & 65.67 & 38.58 & 35.42 \\
    \textsc{BiLSTM}~\cite{DBLP:conf/acl/LiuOSJ20} & 46.94 $\pm$ 1.08 & 46.64 $\pm$ 0.71 & 46.51 $\pm$ 1.11 \\
    \textsc{HAN-BiLSTM}~\cite{DBLP:conf/acl/LiuOSJ20} & 51.93 $\pm$ 1.37 & 49.08 $\pm$ 1.52 & 50.00 $\pm$ 1.49 \\
    \textsc{BERT~\cite{DBLP:conf/emnlp/DurmusLC19}} & 57.19 $\pm$ 0.92 & 55.77 $\pm$ 1.05 & 55.98 $\pm$ 0.70 \\
    \textsc{DisCOC}~\cite{DBLP:conf/acl/LiuOSJ20} & 57.90 $\pm$ 0.70 & 59.41 $\pm$ 1.41 & 58.36 $\pm$ 0.52 \\
    
    \hline
    \textsc{Fine-Tuning (Flan-T5-base)}~\cite{DBLP:journals/jmlr/RaffelSRLNMZLL20} & 58.48 $\pm$ 1.04 & 59.57 $\pm$ 0.54 & 58.45 $\pm$ 0.69\\
    \textsc{Fine-Tuning (Flan-T5-base) (Knowledge)}~\cite{DBLP:journals/jmlr/RaffelSRLNMZLL20}  & 60.37 $\pm$ 1.73 & 60.58 $\pm$ 0.21 & 60.10 $\pm$ 1.25\\
    \textsc{Fine-Tuning (Flan-T5-large)}~\cite{DBLP:journals/jmlr/RaffelSRLNMZLL20} & 58.95 $\pm$ 0.81 & 63.42 $\pm$ 0.52 & 60.23 $\pm$ 0.23 \\
    \textsc{Fine-Tuning (Flan-T5-large) (Knowledge)}~\cite{DBLP:journals/jmlr/RaffelSRLNMZLL20} & 62.63 $\pm$ 1.04 & 64.56 $\pm$ 0.56 & 63.31 $\pm$ 0.68\\ 

    \textsc{Prompt-Tuning (Flan-T5-base)}~\cite{DBLP:conf/emnlp/LesterAC21} & 61.05 $\pm$ 1.58 & 57.80 $\pm$ 0.76 & 58.61 $\pm$ 0.84\\
    \textsc{Prompt-Tuning (Flan-T5-base) (Knowledge)}~\cite{DBLP:conf/emnlp/LesterAC21} & 60.52 $\pm$ 0.32 & 59.78 $\pm$ 0.62 & 59.58 $\pm$ 0.69\\
    \textsc{Prompt-Tuning (Flan-T5-large)}~\cite{DBLP:conf/emnlp/LesterAC21} & 63.48 $\pm$ 1.33 & 63.13 $\pm$ 0.78 & 63.10 $\pm$ 0.77\\  

    \textsc{Prompt-Tuning (Flan-T5-large)(Knowledge)}~\cite{DBLP:conf/emnlp/LesterAC21} & 65.18 $\pm$ 1.03 &  \underline{65.12} $\pm$ 0.75 & 65.20 $\pm$ 0.46\\
    \textsc{Prefix-Tuning (Flan-T5-base)}~\cite{DBLP:conf/acl/LiL20} & 61.95 $\pm$ 2.03 & 57.69 $\pm$ 1.44 & 58.35 $\pm$ 1.24\\
    \textsc{Prefix-Tuning (Flan-T5-base) (Knowledge)}~\cite{DBLP:conf/acl/LiL20} & 61.85 $\pm$ 1.43 & 60.96 $\pm$ 0.17 & 60.13 $\pm$ 0.22\\
    \textsc{Prefix-Tuning (Flan-T5-large)}~\cite{DBLP:conf/acl/LiL20} & 65.01 $\pm$ 0.95 & 62.10 $\pm$ 1.25 & 62.87 $\pm$ 1.01\\

    \textsc{Prefix-Tuning (Flan-T5-large) (Knowledge)}~\cite{DBLP:conf/acl/LiL20} & \underline{65.43} $\pm$ 1.53 & 65.09 $\pm$ 0.64 & \underline{65.25} $\pm$ 0.71\\
    \hline
    
    \textsc{PersonaPrompt (Flan-T5-base)} & 59.19 $\pm$ 0.57 & 60.20 $\pm$ 0.67 & 59.29 $\pm$ 0.67\\
    \textsc{PersonaPrompt (Flan-T5-base) (Knowledge)} & 64.35 $\pm$ 0.63 & 62.11 $\pm$ 0.36 & 62.81 $\pm$ 0.31\\
    \textsc{PersonaPrompt (Flan-T5-large)} & 65.40 $\pm$ 0.54 &64.26 $\pm$ 1.08 &  64.35 $\pm$ 0.51\\ 

    \textsc{PersonaPrompt (Flan-T5-large) (Knowledge)} & \bf{68.48} $\pm$ 1.24 & \bf{67.16} $\pm$ 0.58 &  \bf{67.77} $\pm$ 0.54 \\  
    \bottomrule
    \end{tabular}}
    \vspace{-0.3cm}
    \caption{The mean and standard deviation of the performance of different models on argument impact classification task (i.e., \textit{Kialo} dataset). \textsc{Knowledge} indicates incorporating with the generated audience persona knowledge. The upper portions are the previous SOTA methods, and the middle portions are the implemented baselines.}
    \label{table:arg_impact_main}
    \vspace{-0.6cm}
\end{table*}

\begin{table}[!t]
    \centering
    \small
    \setlength\tabcolsep{2pt}
    \scalebox{0.75}{\begin{tabular}{l|c}
    \toprule
    \multicolumn{1}{c|}{\textbf{Model}} &  \multicolumn{1}{c}{\textbf{Accuracy}} \\ 
    \midrule
    \textsc{Majority} & 62.62 \\
    \textsc{Linguistic+User LR}& 67.41 \\
    \textsc{Arg-Struct LR} & 69.52 \\
    \textsc{Linguistic+Arg-Struct LR} & 70.48 \\
    \textsc{Linguistic+User+Arg-Struct LR} & 70.44 \\
    \hline
    \textsc{BERT} & 64.71 $\pm$ 0.73\\
    \textsc{BERt(Knowledge)} & 66.09 $\pm$ 1.27\\
    \textsc{DisCOC} & 64.48 $\pm$1.03\\
    \textsc{DisCOC(Knowledge)} & 67.01 $\pm$ 1.17\\
    \hline
    \textsc{Fine-Tuning (Flan-T5-base)} & 70.15 $\pm$ 0.58\\
    \textsc{Fine-Tuning (Flan-T5-base)(Knowledge)}  & 72.87 $\pm$ 0.12\\
    \textsc{Fine-Tuning (Flan-T5-large)} & 71.75 $\pm$ 0.59\\
    \textsc{Fine-Tuning (Flan-T5-large)(Knowledge)}  & 73.25 $\pm$ 0.41\\
    \hline
    \textsc{Prefix-Tuning (Flan-T5-base)} & 68.62 $\pm$ 0.77\\
    \textsc{Prefix-Tuning (Flan-T5-base)(Knowledge)} & 71.26 $\pm$ 0.49\\
    \textsc{Prefix-Tuning (Flan-T5-large)} & 70.69 $\pm$ 0.79\\ 
    \textsc{Prefix-Tuning (Flan-T5-large)(Knowledge)} & 73.41 $\pm$ 0.58\\
    \hline
    \textsc{Prompt-Tuning (Flan-T5-base)} & 68.01 $\pm$ 1.38\\
    \textsc{Prompt-Tuning (Flan-T5-base)(Knowledge)} & 71.61 $\pm$ 0.63\\
    \textsc{Prompt-Tuning (Flan-T5-large)} & 71.26  $\pm$ 1.22\\  
    \textsc{Prompt-Tuning (Flan-T5-large)(Knowledge)} & \underline{73.64} $\pm$ 0.49\\
    \hline
    \textsc{PersonaPrompt (Flan-T5-base)} & 71.08 $\pm$ 0.70\\
    \textsc{PersonaPrompt (Flan-T5-base)(Knowledge)} & 73.45 $\pm$ 0.34\\
    \textsc{PersonaPrompt (Flan-T5-large)} & 72.56 $\pm$ 1.06\\
    \textsc{PersonaPrompt (Flan-T5-large)(Knowledge)} &  \bf{75.86} $\pm$ 0.43\\
    \bottomrule
    \end{tabular}}
    \vspace{-0.3 cm}
    \caption{The performance of the models on the argument persuasion prediction task (i.e., \textit{DDO} dataset), where the upper portions of baselines are Logistic Regression (LR) models with the linguistic feature, user information, and argument structure~\cite{DBLP:conf/emnlp/LiDC20}.}
    \label{table:DDO_main}
    \vspace{-0.6cm}
\end{table}

\section{Experimental Result}
\subsection{Main Results}
Table~\ref{table:arg_impact_main} and Table~\ref{table:DDO_main} summarize the main results of the two online debate tasks, which include the argument impact classification task and argument persuasion task, from which we derive the following conclusions.
\textbf{First}, our method significantly outperforms all baselines in both tasks and achieves state-of-the-art (SOTA) performance in the argument impact classification task. Specifically, our method (Flan-T5-large and Flan-T5-base) outperforms previous SOTA \textsc{DisCOC}~\cite{DBLP:conf/acl/LiuOSJ20} with at least 9.41\% and 4.45\% F1 scores in the argument impact classification task.
\textbf{Second}, our model gains a considerable improvement of 7.54\% F1 score and 4.11\% accuracy over the fine-tuning of the Flan-T5-large (without persona knowledge) model in the \textit{Kialo} and \textit{DDO} datasets. It demonstrates that our method effectively utilizes the audience persona knowledge, perceives this specific knowledge on the correlation of knowledge and the data argument and context, and finally enhances the ability of Flan-T5 to undertake this challenging task.
\textbf{Third}, all models fine-tuned or prompt-tuned with the knowledge exhibit improvement over original tuning. In particular, PersonPrompt (knowledge) with the Flan-T5-base version obtained a 3.52\% F1 score gain in performance over original tuning without persona knowledge. It illustrates the effectiveness of the generated persona knowledge on the online debate quality assessment tasks.

\begin{table}[!t]
\centering
\scalebox{0.6}{\begin{tabular}{l|l|l|l}
\toprule
\multicolumn{1}{c|}{\textbf{Model}} &  \multicolumn{1}{c|}{\textbf{Precision}} &  \multicolumn{1}{c|}{\textbf{Recall}} &  \multicolumn{1}{c}{\textbf{F1}} \\ 
\hline
\textsc{Majority} & 19.43 & 33.33 & 24.55 \\
\textsc{ChatGPT (baseline)} & 40.20 & 33.84 & 34.26 \\
\textsc{ChatGPT (w knowledge)} & 39.04 & 37.18 & 36.60 \\
\textsc{GPT-4 (baseline)} & 50.00 & 44.84 & 39.52 \\
\textsc{GPT-4 (w knowledge)} & 56.14 & 44.19 & 41.60 \\
\hline
\textsc{ChatGPT (A.)} & 47.30 & 33.98 & 26.46  \\ 
\textsc{ChatGPT (A. \& K.)} & 40.55 & 39.03 & 28.44 \\
\textsc{ChatGPT (C. \& A.)} & 41.20 & 34.84 & 35.26 \\
\textsc{ChatGPT (C. \& A.\& K.)} & 39.04 & 37.18 & 36.60  \\
\hline
\end{tabular}}
\vspace{-0.3cm}
\caption{The performance of ChatGPT (\textit{gpt-3.5-turbo-0125}) and GPT-4 (\textit{gpt-4-turbo-2024-04-09}) on the argument impact task. The bottom part is the ChatGPT model performance for the ablation study on the data input. \textsc{C.}, \textsc{A.}, and \textsc{K.} present context, argument, and persona knowledge, respectively.
}
\label{tab:ChatGPT_Performance}
\vspace{-0.6cm}
\end{table}

\subsection{Knowledge Adaptation on Large Language Models}
With the remarkable ability demonstrated by LLMs across a diverse array of tasks~\cite{DBLP:journals/corr/abs-2303-12712,DBLP:journals/corr/abs-2302-04023,DBLP:conf/eacl/ChanCWJFLS24}, we are intrigued about the capability of large language models on zero-shot online debate tasks. We employ the prompting template in \citet{DBLP:conf/iclr/RobinsonW23} to formulate the task as a multiple choice question answering problem as a baseline and append with the audience persona knowledge to compare with this baseline. 
The prompting template is displayed in Figure~\ref{fig:ChatGPT_prompt_tempalte} in Appendix~\ref{sec:Implementation_Details_of_LLMs}.
We test the performance of ChatGPT (\textit{gpt-3.5-turbo-0125})~\cite{openai2022chatgpt} and GPT-4 (\textit{gpt-4-turbo-2024-04-09})~\cite{DBLP:journals/corr/abs-2303-08774} on the argument impact classification task, and the performance is presented in Table~\ref{tab:ChatGPT_Performance}. Although all designed templates perform better than the majority baseline, their overall performance remains suboptimal compared to supervised learning. This result reveals that argument impact classification is still tricky for ChatGPT and cannot be solved easily at the current state, resulting from the argument quality assessment task requiring more ability than only comprehending the semantic meaning of the arguments presented in a debate. As shown in Table~\ref{tab:ChatGPT_Performance}, the context plays a significant role in ChatGPT's zero-shot performance in this task. Moreover, we observed a slight improvement in performance after concatenating knowledge with the pre-designed templates and demonstrated that persona knowledge is also effective for ChatGPT.

\subsection{Ablation Study on PersonaPrompt}
To better investigate the factors of PersonaPrompt, we design numerous ablations on the various aspects of PersonaPrompt on argument impact classification task.
\begin{table}[!t]
    \centering
    \small
    \setlength\tabcolsep{1.5pt}
    \scalebox{0.75}{\begin{tabular}{l|c|c|c}
    \toprule
    \multicolumn{1}{c|}{\textbf{Model}} &  \multicolumn{1}{c}{\textbf{Precision}} &  \multicolumn{1}{c}{\textbf{Recall}} &  \multicolumn{1}{c}{\textbf{F1}} \\ 
    \midrule
    \textsc{PersonaPrompt (A.)} & 56.94 & 56.08 & 56.28\\
    \textsc{PersonaPrompt (A. \& C.)} & 59.19 & 60.20 & 59.29\\
    \textsc{PersonaPrompt (A. \& K.)} & 60.52 & 60.15 & 60.23\\
    \textsc{PersonaPrompt (A. \& C. \& K.)} & \textbf{64.35} & \textbf{62.11} & \textbf{62.81}\\
    
    \hline
    
    \textsc{BERT (A.)} &  53.24 & 50.93 & 51.53 \\
    \textsc{BERT (A. \& C.)} & 57.19 & 55.77 & 55.98 \\
    \textsc{BERT (A. \& K.)} & 53.52 & 54.94 & 53.59 \\
    \textsc{BERT (A. \& C. \& K.)} & 56.76 & 58.55  & 57.25 \\
    \textsc{DisCOC (A. \& C.)} & 57.90 & 59.41 & 58.36 \\
    \textsc{DisCOC (A. \& C. \& K.)} & 57.83 & 59.94 & 58.69 \\

    \textsc{Flan-T5 (A.)} & 49.26 & 54.99 & 50.44\\
    \textsc{Flan-T5 (A. \& C.)} & 58.48 & 59.57 & 58.45 \\
    \textsc{Flan-T5 (A. \& K.)} & 54.39 & 58.16 & 55.62\\
    \textsc{Flan-T5 (A. \& C. \& K.)} & 60.37 & 60.58 & 60.10\\
    \bottomrule
    \end{tabular}
    }
    \vspace{-0.3cm}
    \caption{The ablation study on the argument impact classification task, where  \textsc{C.}, \textsc{A.}, and \textsc{K.} stands for context, argument, and persona knowledge, respectively. Note that DisCOC must require the context due to its recurrent mechanism. Flan-T5 indicates the Fine-Tuning (Flan-T5-base) model.
    }
    \label{table:arg_impact_ablationstudy}
    \vspace{-0.5cm}
\end{table}

\paragraph{Can Knowledge Replace Context}
The context has demonstrated significant influences on the model performance in prior works~\cite{DBLP:conf/emnlp/DurmusLC19, DBLP:conf/acl/LiuOSJ20}, so we wonder whether persona knowledge can be used to replace the context.
As the experimental results reported in Table~\ref{table:arg_impact_ablationstudy}, we draw the following conclusions:
(1) Context significantly contributes to the performance of various models;
(2) Persona knowledge can be utilized to replace certain information or signals from the context, aiding the model in determining the class label and emphasizing the importance of knowledge;
(3) Incorporating persona knowledge alongside the context consistently improves model performance.

\begin{figure}[!t]
    \centering
    \includegraphics[width=0.8\linewidth]{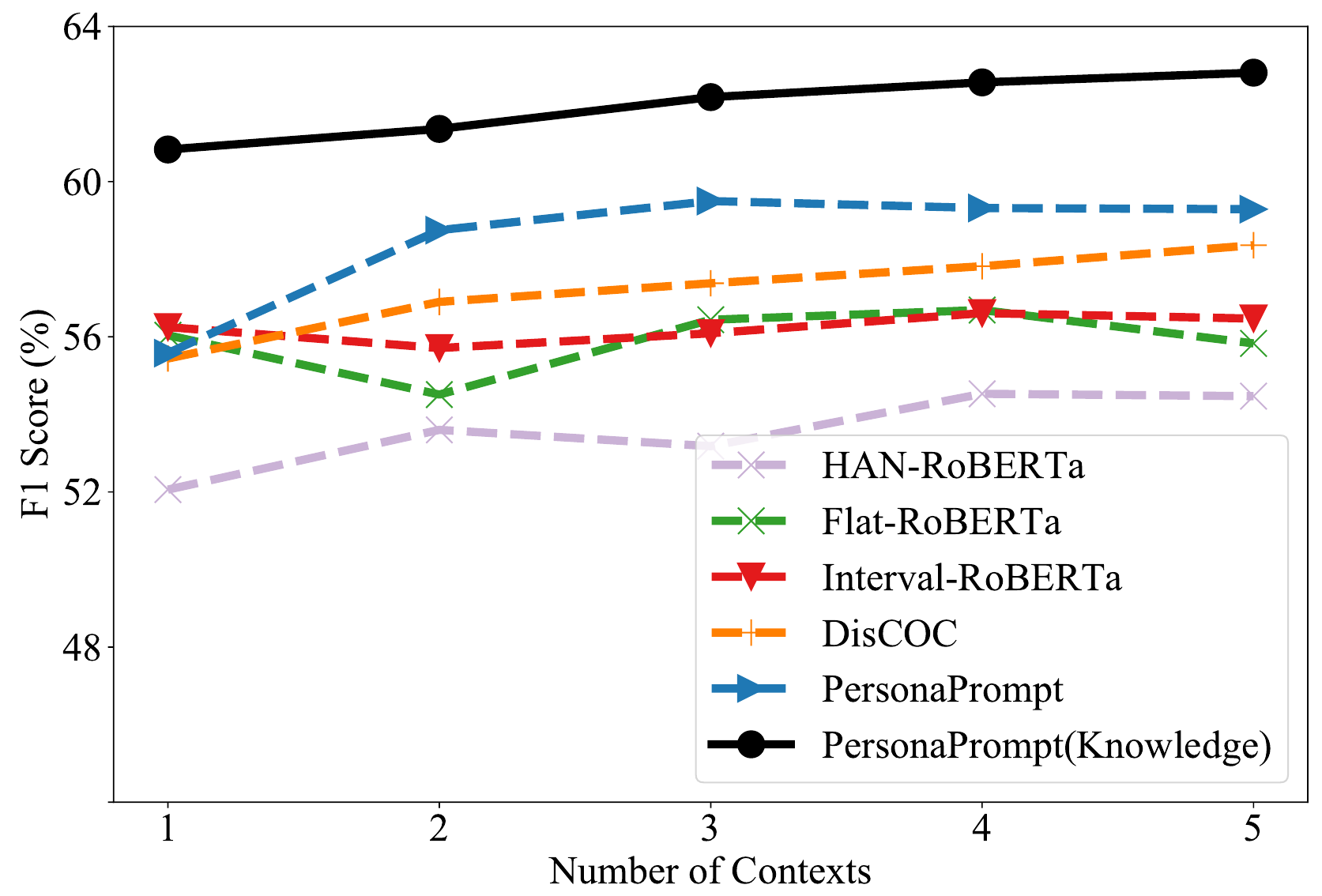}
    \vspace{-0.4cm}
    \caption{F1 scores of different models on varying the context numbers. The results of HAN, Flat, and Interval-RoBERTa are referenced from~\citet{DBLP:conf/acl/LiuOSJ20}. The distinguishing factor among these models lies in the form of context modeling.}\label{fig:path_len}
    \vspace{-0.4cm}
\end{figure}

\paragraph{Influence of the Context Length}
Different debate claims have different context lengths in the Kailo dataset. 
Figure~\ref{fig:path_len} shows F1 scores of models with different context lengths. 
Only \textsc{PresonaPrompt (Knowledge)} and \textsc{DisCOC} benefit from longer discourse contexts, while other models get stuck in performance fluctuation. \textsc{PresonaPrompt (Knowledge)} and \textsc{DisCOC} have consistent performance gains; instead, other models cannot learn long-distance structures better. With the persona knowledge, the PLMs can receive extra signals to perceive the semantics meanings of longer context.

\paragraph{Prompt Engineering} Furthermore, we conduct the prompt template searching and the parameter sensitivity on the continuous prompt length that we describe in Appendix~\ref{sec: Appendix_ablation_study_PersonaPrompt}.

\subsection{In-depth Exploring on Persona Knowledge}

\paragraph{Are All Persona Dimensions Helpful}
Experiments are conducted on all four dimensions of persona knowledge to verify the effectiveness of each dimension. Based on the result presented in Table~\ref{table:arg_impact_dimension_ablationstudy}, it can be concluded that the persona argument is the most essential dimension of the persona that contributes to the model performance. For instance, as demonstrated in Figure~\ref{fig:DataInstance_Introduction}, the computer science professor persona instantiates low-level concepts such as memory management, hardware interactions, and performance optimization in the argument, thereby providing extensive information to strengthen their viewpoint on the debate topic for pre-trained language models to undertake this task. Moreover, other dimensions also provide additional signals to help the pre-trained language models determine the impact and persuasiveness of the argument.

\begin{table}[!t]
    \centering
    \small
    \scalebox{0.65}{
    \begin{tabular}{l|c c c}
    \toprule
    \multicolumn{1}{c|}{\textbf{Model}} &  \multicolumn{1}{c}{\textbf{Precision}} &  \multicolumn{1}{c}{\textbf{Recall}} &  \multicolumn{1}{c}{\textbf{F1}} \\ 
    \midrule
    \textsc{PersonaPrompt (w/o knowledge)}  & 59.19 $\pm$ 0.57 & 60.20 $\pm$ 0.67 & 59.29 $\pm$ 0.67\\
    \textsc{PersonaPrompt (w Knowledge)} & \bf{64.35} $\pm$ 0.63 & \bf{62.11} $\pm$ 0.36 & \bf{62.81} $\pm$ 0.31\\
    \hline
    \textsc{PersonaPrompt (R \& S)} & 61.06 $\pm$ 0.86 & 61.38 $\pm$ 1.26 & 60.63 $\pm$ 1.34\\
    \textsc{PersonaPrompt (R \& A)} & 61.54 $\pm$ 1.03 & 61.38 $\pm$ 1.20 & 61.26 $\pm$ 1.61\\
    \textsc{PersonaPrompt (R \& C)} & 61.57 $\pm$ 0.77 & 60.70 $\pm$ 0.62 & 60.85 $\pm$ 0.29\\
    \textsc{PersonaPrompt (R \& I)} & 62.29 $\pm$ 0.76 & 60.65 $\pm$ 0.78 & 60.79 $\pm$ 1.31\\ 
    \textsc{PersonaPrompt (R \& A \& S)} & 62.10 $\pm$ 0.88 & 62.98 $\pm$ 0.81 & 62.09 $\pm$ 1.25\\
    \textsc{PersonaPrompt (R \& A \& C)} & 61.07 $\pm$ 0.89 & 63.10 $\pm$ 0.85 & 61.67 $\pm$ 0.84\\
    \textsc{PersonaPrompt (R \& A \& I)} & 62.77 $\pm$ 1.22 & 62.26 $\pm$ 0.59 & 62.29 $\pm$ 0.12\\
    \textsc{PersonaPrompt (R \& S \& A \& C)} & 62.74  $\pm$ 1.12 & 62.51 $\pm$ 0.82 & 62.44 $\pm$ 0.48\\ 
    \textsc{PersonaPrompt (R \& S \& A \& I)} & 63.97 $\pm$ 0.83 & 61.94 $\pm$ 0.68 & 62.73 $\pm$ 1.13\\  
    \hline
    
    \textsc{PersonaPrompt (1 Persona)} & 61.18 $\pm$ 1.04 & 58.84 $\pm$ 0.79 & 59.58 $\pm$ 1.17\\
    \textsc{PersonaPrompt (2 Personae)} & 61.03 $\pm$ 1.30 & 60.49 $\pm$ 0.64 & 60.60 $\pm$ 1.15\\
    \textsc{PersonaPrompt (3 Personae)} & 61.62 $\pm$ 1.12 & 60.66 $\pm$ 0.92 & 61.05 $\pm$ 0.56\\
    \textsc{PersonaPrompt (4 Personae)} & 62.50 $\pm$ 0.74 & 63.27 $\pm$ 1.06 & 61.98 $\pm$ 0.93\\
    \textsc{PersonaPrompt (5 Personae)} & 63.98 $\pm$ 1.13 & 61.94 $\pm$ 1.01 & 62.67 $\pm$ 1.10\\
    \hline 
    \textsc{PersonaPrompt (Con)} & 60.33 $\pm$ 0.88 & 61.99 $\pm$ 0.89 & 61.05 $\pm$ 0.44 \\
    \textsc{PersonaPrompt (Pro)} & 61.07 $\pm$ 0.90 & 60.73 $\pm$ 0.94 & 60.79 $\pm$ 0.62 \\
    \textsc{PersonaPrompt (Neutral)} & 63.15 $\pm$ 0.87 & 62.05 $\pm$ 0.96 & 61.95 $\pm$ 0.75\\
    \bottomrule
    \end{tabular}
    }
    \vspace{-0.2cm}
    \caption{The ablation study on persona knowledge on the argument impact classification task. \textsc{R}, \textsc{S}, \textsc{A}, \textsc{C}, and \textsc{I} represent role, stance, argument, character, and intent.
    }
    \label{table:arg_impact_dimension_ablationstudy}
    \vspace{-0.6 cm}
\end{table}

\paragraph{Influence of Persona Number}
To obtain a more profound comprehension of the effect of persona number on the model performance, a series of experiments are designed with varying quantities of persona, with results illustrated in Table~\ref{table:arg_impact_dimension_ablationstudy}. Generally, it is observed that an increase in the number of personae leads to a corresponding increase in the model performance, with noteworthy enhancement in the 3.09\% F1 score observed when five personae were employed as opposed to 1 persona. However, the optimal PersonaPrompt (persona knowledge) equipped with more than five personas does not observe significant benefits, and it may result from the limited maximum sequence length of model input.

\paragraph{Are Stance Group Helpful}
To probe a deeper understanding of the stance group of persona (i.e., \textsc{Pro, Con}, and \textsc{Neutral}) contributed to the performance, we divided the persona into three distinct groups and performed experiments with the same quantity of persona in each group. To ensure a fair comparison, we opt for all the persona knowledge within these three groups that possess similar token lengths.
Table~\ref{table:arg_impact_dimension_ablationstudy} reveals an intriguing finding that the \textsc{neutral} group outperforms the other groups. 
Specifically, the P-values for the \textsc{Neutral} group are 0.0286 and 0.0493 (paired student’s t-test, p < 0.05) against the \textsc{PRO} and \textsc{CON} groups. One rationale may be the \textsc{Neutral} group persona knowledge encoding more information or signal on both sides instead of just a single side and their stances, as illustrated by the example depicted in Figure~\ref{fig:DataInstance_Introduction}.

\begin{table}[!t]
\small
\centering
\scalebox{0.75}{\begin{tabular}{l|l|l|l}
\toprule
\multicolumn{1}{c|}{\textbf{Model}} &  \multicolumn{1}{c|}{\textbf{Precision}} &  \multicolumn{1}{c|}{\textbf{Recall}} &  \multicolumn{1}{c}{\textbf{F1}} \\ 
\hline
\textsc{w/o knowledge}& 59.19 $\pm$ 0.57 & 60.20 $\pm$ 0.67 & 59.29 $\pm$ 0.67\\
\textsc{ConceptNet (Triple)}& 60.20 $\pm$ 0.80 & 59.59 $\pm$ 1.13 & 59.46 $\pm$ 0.35 \\
\textsc{ConceptNet (Language)} & 60.09 $\pm$ 0.82 & 61.08 $\pm$ 0.78 & 59.89 $\pm$ 0.75 \\
\textsc{Background knowledge} & 61.21 $\pm$ 1.22 & 60.59 $\pm$ 0.98 & 60.09 $\pm$ 1.10 \\
\textsc{Persona knowledge} & 64.35 $\pm$ 0.63 & 62.11 $\pm$ 0.36 & 62.81 $\pm$ 0.31\\
\hline
\end{tabular}}
\vspace{-0.4cm}
\caption{The performance of PersonaPrompt (Flan-T5-base) with various knowledge resources on the argument impact task. \textsc{Triple} and \textsc{Language} correspond to the ConceptNet knowledge representation forms in triple and natural language, respectively.
}
\label{tab:KnowledegSource_Performance}
\vspace{-0.5cm}
\end{table}

\begin{figure*}[!t]
    \centering
    \includegraphics[width=0.7\linewidth]{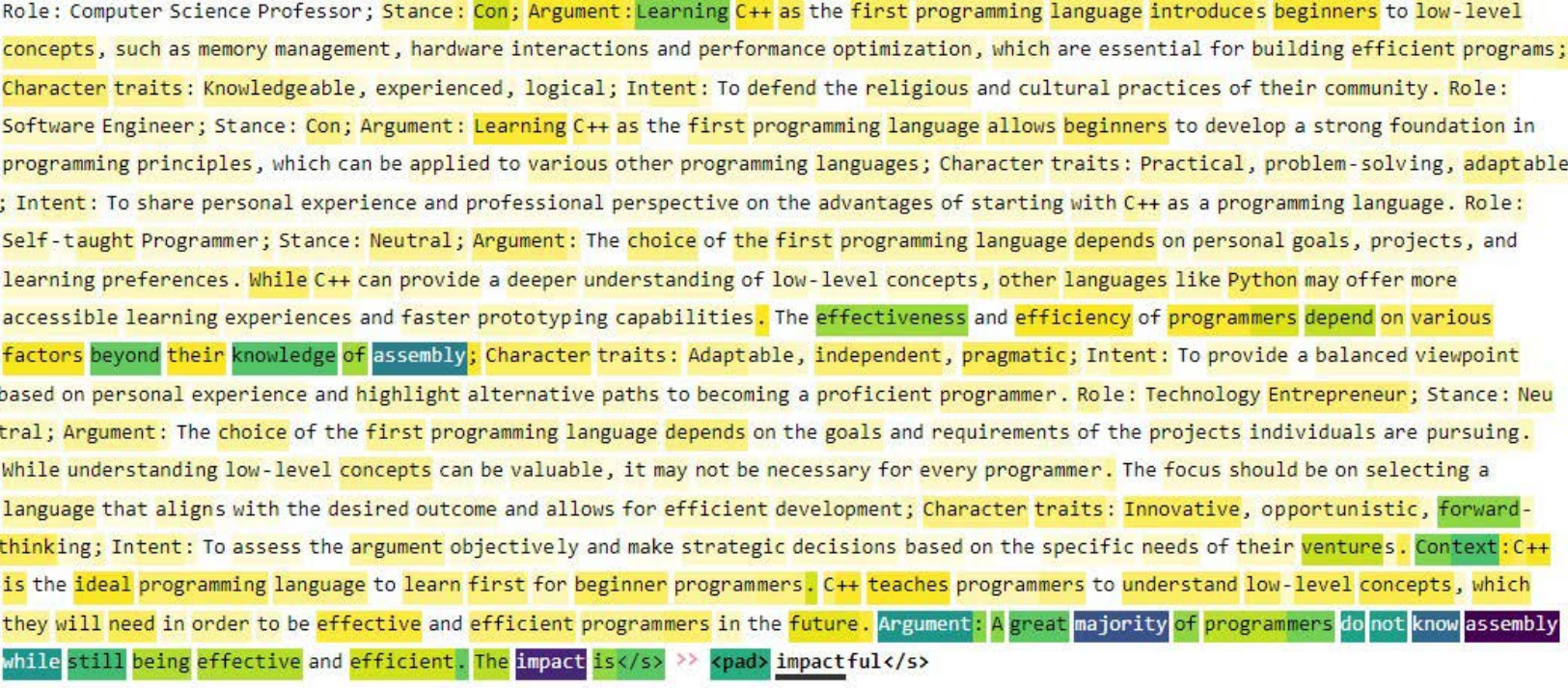}
    \vspace{-0.3cm}
    \caption{Attention visualization for \textsc{Fine-Tuning (FLAN-T5-base) (Knowledge)} on the example shown in Figure~\ref{fig:DataInstance_Introduction}.
    }\label{fig:Attention_Visualization}
    \vspace{-0.7cm}
\end{figure*}

\subsection{Knowledge Type Comparison}
We undertake a comparison of the generated persona knowledge with the commonsense knowledge from ConceptNet and background knowledge generated from ChatGPT to assess the exact contribution of persona knowledge and the effectiveness of different knowledge on this argument impact classification task. ConceptNet~\cite{DBLP:conf/aaai/SpeerCH17} is a widely used and traditional knowledge graph consisting of 42 relation types. By following the \textsc{KagNet} method~\cite{DBLP:conf/emnlp/LinCCR19}, we ground the ConceptNet knowledge on the argument and context of the Kailo dataset. The case example of retrieved commonsense knowledge can be found in Figure~\ref{fig:ConceptNet_Knowledge} and Figure~\ref{fig:ConceptNet_Knowledge_1} in Appendix~\ref{sec:More_case_ConceptNet}. Furthermore, we crafted a prompt template, "Please list all relevant background knowledge regarding the argument and context," to generate the background knowledge from ChatGPT. After receiving all retrieved knowledge, we substitute the persona knowledge with the commonsense or background knowledge in our designed input template shown in Figure~\ref{fig:DiscoursePrompt_Architecture}. There are two representations of commonsense knowledge from ConceptNet, which are the triple and natural language representing forms. 
The outcome is reported in Table~\ref{tab:KnowledegSource_Performance} and indicates that the ConceptNet knowledge does not make a significant improvement on this task. The reason behind this result may be the retrieval of much noisy and contextually irrelevant knowledge from the traditional knowledge graph that damages the model performance. This problem remains a challenging research question~\cite{DBLP:conf/emnlp/LinCCR19}, while our generated persona knowledge from ChatGPT obtained a high human evaluation score on the relevance aspect. Moreover, the persona knowledge model obtains a significant performance gap against the background knowledge, and it evidences the efficacy of multi-dimensional persona knowledge.

\subsection{Attention Visualization}
To further examine the impact of persona knowledge on this argument impact classification task, we display how the model (i.e., \textsc{Fine-Tuning (Flan-T5-base) (Knowledge)}) assigns weight to different input elements by using an attention visualization tool~\cite{alammar2020explaining}, and the resulting visualization is shown in Figure~\ref{fig:Attention_Visualization}. Interestingly, the Self-taught Programmer contributes the highest weights among all persona roles and even surpasses the weight contributed by the context. A neutral persona and their argument seem to provide more information and weights to assist PLMs in determining the class label, which is consistent with previous findings. More cases of attention visualization can be found in Appendix~\ref{sec:More_case_Attention_Visualization}.


%% file: sections/related_work.tex
\section{Related Work}
\subsection{Argument Persuasiveness Classification}
The study of computational argumentation has recently attracted more attention, which uses corpora collected from web argumentation sources like the CMV sub-forum of Reddit to assess the qualitative impact of arguments~\cite{DBLP:conf/www/TanNDL16}. There are many literature studies on the importance and efficacy of various aspects in determining the persuasiveness, including surface textual, social interaction, and argumentation-related features~\cite{DBLP:conf/acl/WeiLL16}, the characteristics of the source~\cite{DBLP:conf/www/DurmusC19} and audience~\cite{DBLP:conf/naacl/DurmusC18,DBLP:conf/um/Shmueli-Scheuer19}, the sequence ordering of argument~\cite{DBLP:conf/aaai/HideyM18}, and style feature aspects~\cite{DBLP:conf/acl/BaffWKS20}, were studied and investigated. Apart from the aforementioned features, 
\citet{DBLP:conf/emnlp/DurmusLC19} turned to the pragmatics and discourse context in the analysis of arguments.
They conducted experiments to demonstrate that the historical arguments are beneficial for the model performance to some extent. \citet{DBLP:conf/www/Zeng0HGLK20} and \citet{DBLP:conf/acl/LiuOSJ20} performed research on how the context and dynamic progress of argumentative conversation affect comparative persuasiveness in the debate process.

\subsection{Knowledge Elicitation from Pre-trained Language Models}
Considerable studies have demonstrated that Pre-trained Language Models (PLMs) have a substantial amount of knowledge implicitly stored that can be accessed via conditional generation~\cite{DBLP:conf/emnlp/PetroniRRLBWM19,DBLP:conf/emnlp/DavisonFR19, DBLP:journals/tacl/JiangXAN20}.
Additionally, the giant GPT-3~\cite{brown2020language} showed that manually designed prompts can tailor generations for diverse tasks in few-shot scenarios and achieve competitive results.
Accordingly, prompt tuning methods can use these language models to directly elicit knowledge to perform language understanding~\cite{DBLP:conf/emnlp/WangHQSWLG22,DBLP:journals/corr/abs-2305-05252} and commonsense reasoning~\cite{DBLP:conf/coling/YangLNTWL20,DBLP:conf/acl/ParanjapeMGHZ21,DBLP:conf/acl/0010LLWWBCH22}.
Recently, ChatGPT has demonstrated its ability to assume various roles and perform tasks in different domains based on given instructions~\cite{DBLP:journals/corr/abs-2305-14325, DBLP:journals/corr/abs-2305-14688, DBLP:journals/corr/abs-2305-02364, DBLP:conf/emnlp/JiangCCW23, DBLP:journals/corr/abs-2408-02559}, especially simulating an open world~\cite{DBLP:journals/corr/abs-2304-03442}.
Therefore, we employ ChatGPT to emulate the audience of diverse backgrounds in debates and utilize a prompt to inject persona knowledge into classifiers.

%% file: sections/conclusion.tex
\section{Conclusion}
This paper introduces a persona knowledge-aligned prompt tuning method for tackling online debate argument tasks by utilizing audience persona knowledge. Our proposed framework elicits this persona knowledge from a large language model (i.e., ChatGPT).  
The performance of our model exhibits significant and consistent improvement against competitive baselines. We hope our comprehensive discussions will provide valuable insights for communities in computational argumentation.

%% file: sections/limitation.tex
\section*{Limitations}
\paragraph{Limited Persona Knowledge Dimensions}
First, the generated persona knowledge cannot be deemed comprehensive. Although we select four dimensions of personas based on the task, the generated persona knowledge cannot cover all dimensions of personas, nor all attributes of these dimensions.
Furthermore, since our generated persona knowledge only utilizes the context and argument from the dataset, the knowledge is task-specific on the online debate argument tasks, including the argument impact classification task and the argument persuasion prediction task.

\paragraph{Reporting biases Inherent in Pre-trained Language Model}
All elicited persona knowledge only comes from ChatGPT and its pre-training corpora. This constraint restricts the persona knowledge types, numbers, and role types for each debate topic owing to the reporting biases~\cite{gordon2013reporting} in the pre-trained language models (PLMs). This constraint consequently restricts the model capability in the online debate tasks.

%% file: sections/ethics_statement.tex
\section*{Ethics Statement}
In this work, we conformed to recognized privacy practices and rigorously followed the data usage policy. We declare that all authors of this paper acknowledge the \emph{ACM Code of Ethics} and honor the code of conduct. 
This paper presents a method to utilize generated audience persona knowledge from ChatGPT to provide more signals to enhance the model performance on two online debate tasks. All generated persona knowledge reflects the selection and reporting biases~\cite{gordon2013reporting} of ChatGPT, which could sometimes be stereotypical and do not represent the views of the authors. However, we took the following steps to mitigate this effect. Firstly, we design an explicit prompt to instruct the ChatGPT to generate optimistic attributes about personas, which has been shown in prior work to reduce the toxicity of outputs~\cite{DBLP:journals/tacl/SchickUS21}. Second, we performed the human evaluations on the 1,000 sampled persona knowledge generated from ChatGPT and did not observe any harmful and toxic language resulting from the ChatGPT fine-tuned with the RLHF approach~\cite{DBLP:conf/nips/ChristianoLBMLA17}, which prevents ChatGPT from generating harmful and toxic language without deliberate attacks~\cite{DBLP:conf/nips/ChristianoLBMLA17}. Nevertheless, it is essential to acknowledge that none of these safeguards are perfect. We cannot guarantee that all generated persona knowledge does not contain any undesired or harmful content, and expert annotators may possess varying perspectives on what constitutes toxic content~\cite{DBLP:conf/naacl/SapSVZCS22}.

%% file: sections/appendix.tex
\appendix
\section{PersonaPrompt}
\subsection{PresonaPrompt Implementation Details} \label{sec:appendix_discoPrompt_implementation_details}
PresonaPrompt is prompt tuning upon Flan-T5-model, and we also validate our method over two model scales, including Flan-T5-base and Flan-T5-large. Generally, the overall configuration follows the setting in \citet{DBLP:conf/emnlp/LesterAC21} and sets the learnable prompt length as 20. For the argument impact classification task, the training was implemented using cross-entropy loss with 30,000 training steps, which selects the model that yields the best performance on the validation set. We adopt an Adafactor~\cite{DBLP:conf/icml/ShazeerS18} optimizer with various learning rate ranges for different dataset settings. The batch size and maximum input sequence are 4 and 512, respectively. The maximum generated sequence length of the encoder is 10. Our model is conducted on two 32GB NVIDIA V100 GPUs. The running time for Flan-T5-base is around 8 hours, while Flan-T5-large is about 22 hours. 
The tailored prompt template is shown in Figure~\ref{fig:PresonaPrompt_Template_Searching}. The specific hyperparameters of implementation details for PresonaPrompt(Flan-T5-base) and PresonaPrompt(Flan-T5-large) are displayed in Table~\ref{table:Implementation_details_hypreparameters}. The frozen pre-trained Flan-T5 model download from \textit{HuggingFace}, and our model inheritance and modification from \textsc{OpenPrompt}~\cite{DBLP:conf/acl/DingHZCLZS22}. 

\subsection{A Persona Knowledge Generated by the GPT-4}\label{sec:More_case_GPT-4}
\begin{figure*}[t]
    \centering
    \includegraphics[width=0.7\linewidth]{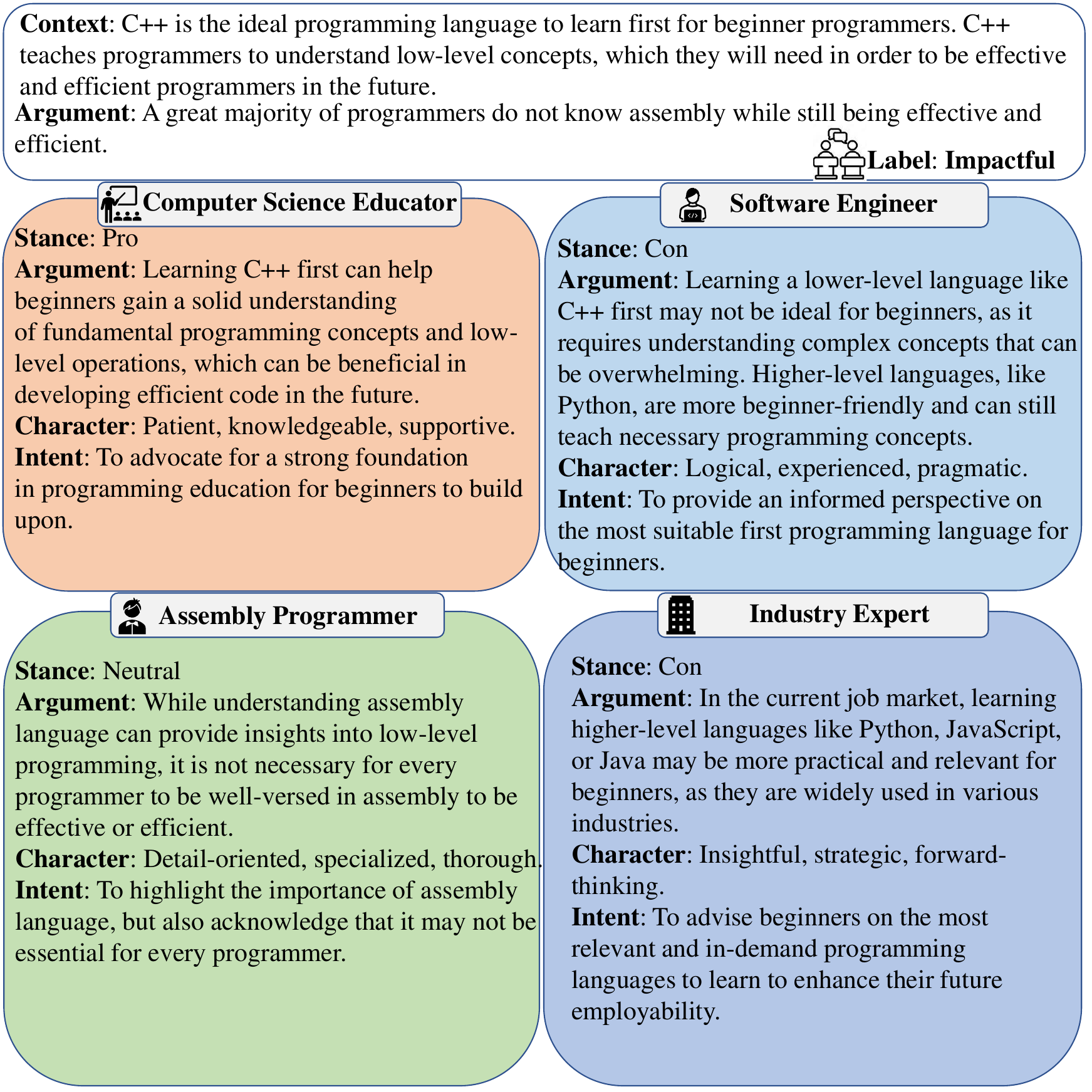}
    \caption{Persona knowledge generated by GPT-4 on the example shown in Figure 1 in the main paper context.}
    \label{fig:Persona_Knowledge_GPT4}
\end{figure*}

A Persona knowledge generated by the GPT-4 is displayed in Figure~\ref{fig:Persona_Knowledge_GPT4}. The application of GPT-4 to produce all persona knowledge is difficult due to the constraints imposed by its usage quota and cost of GPT-4. It is noteworthy that the persona knowledge generated by ChatGPT and GPT-4 exhibits great similarity. As a result,  we have opted for ChatGPT to generate all persona knowledge.

\section{Appendix for Experimental Settings}\label{sec:appendix_experiments}

\begin{table}[t]
\small
\centering
\begin{tabular}{l|c}
\toprule
\textbf{Dataset} & \textbf{Hyperparameters} \\

\midrule
\multirow{2}{*}{Kialo}          & LR$^{*}$: 3e-6,\\
                                & BS: 4, gradient accumulation step:1\\
\midrule
\multirow{2}{*}{DDO}            & LR$^{*}$: 3e-7,\\
                                & BS: 4, gradient accumulation step:1\\
\midrule
\midrule
\multirow{2}{*}{Kialo}          & LR$^{*}$: 4e-6,\\
                                & BS: 4, gradient accumulation step:1 \\
\midrule
\multirow{2}{*}{DDO}            & LR$^{*}$: 3e-7,\\
                                & BS: 4, gradient accumulation step:1\\
\midrule                               
\bottomrule 
\end{tabular}
\caption{The hyperparameters of implementation details for PersonaPrompt (Flan-T5-base) and PersonaPrompt (Flan-T5-large). The upper part is for the Flan-T5-base version in two datasets, while the bottom is for the Flan-T5-large version. ``LR$^{*}$'' and ``BS'' refer to optimal learning rate and batch size, respectively. We perform grid search with learning rates \{3e-6, 4e-6, 5e-6\} in the Kialo dataset and grid search with learning rates \{3e-7, 4e-7, 5e-7\} in the DDO dataset.}

\label{table:Implementation_details_hypreparameters}
\end{table}

\subsection{DataSet} \label{sec:appendix_data_statistics}
\paragraph{Kialo} We conduct the experiments on the argument impact classification task from the \textit{kialo} dataset and follow the experimental setting of the previous studies \cite{DBLP:conf/emnlp/DurmusLC19, DBLP:conf/acl/LiuOSJ20}. The statistics of this dataset are presented in Table~\ref{table:stat_label}, and this dataset contains 47,219 argument claims from \textit{kialo.com} for 741 controversial theme and their corresponding impact votes. Following the setting of the previous studies \cite{DBLP:conf/emnlp/DurmusLC19, DBLP:conf/acl/LiuOSJ20}, the macro F1 score is employed as our main evaluation metric while the Precision and Recall metrics are taken into account in our work. The impact vote for each argument claim is provided by users who can choose from 1 to 5, and \citet{DBLP:conf/emnlp/DurmusLC19} categorize votes into three impact classes (\textit{Not Impactful, Medium Impact, and Impactful}) based on the agreement and the valid vote numbers to reduce noise. The label definition of debate platform is in ~\url{https://www.kialo.com/tour}.
\begin{table}[!t]
    \small
    \centering
    \scalebox{1}{\begin{tabular}{c|c|c|c}
        \toprule
         \textbf{Impact} & \textbf{Train} & \textbf{Validation} & \textbf{Test} \\
        \hline
        \textit{Impactful} & 3,021 & 641 &  646 \\
        \textit{Medium Impact} & 1,023  & 215 & 207 \\
        \textit{Not Impactful} & 1,126 & 252 & 255 \\
        \hline
        \textit{Total} & 5,170 & 1,108 & 1,108 \\        
        \bottomrule
    \end{tabular}}
    \caption{Data statistics of \textit{Kialo} dataset.}
    \label{table:stat_label}
\end{table}

\paragraph{DDO} We conduct the experiment with the DDO dataset~\cite{DBLP:conf/acl/DurmusC19}, which comprises 77,655 debates that cover 23 different topic categories. Each debate consists of multiple rounds, each containing one utterance from the PRO side and one from the CON side. In addition to the text information for debates, the dataset also collects user information and votes provided by the audience on six different criteria for evaluating both the debaters. In the argument persuasion prediction task, the DDO dataset~\cite{DBLP:conf/acl/DurmusC19, DBLP:conf/emnlp/LiDC20} was employed to predict the winner who made more convincing arguments among the two debaters. 
By following~\citet{DBLP:conf/emnlp/LiDC20}, debates were eliminated if they were tied or the difference in votes was only one. Furthermore, all debates with more than 40 sentences in one round from one side were excluded since the model takes a relatively long inference time and performs worse for long debates. Debates where one of the debaters forfeited during the discussion were also removed. Ultimately, the final dataset contained 2,608 debates, split into 870, 869, and 869 as three folds.

\subsection{Baseline Models}
\label{sec:appendix_baseline_models}
To exhibit the effectiveness of our proposed method, we compared it with previous works on the Kialo and DDO datasets. In this section, we mainly describe some listed baselines in Table 2 in the main paper context, and more baselines can be found in Table~\ref{table:arg_impact_supple}.

\begin{itemize}[leftmargin=*]
    \item \textbf{Majority.} The baseline simply returns majority class in dataset. For example, in kialo, the baseline simply predict label \textit{Impactful}.

    \item \textbf{SVM.}~\citet{DBLP:conf/emnlp/DurmusLC19} created linguistic features for a SVM classifier, such as named entity types, POS tags, special marks, tf-idf scores for n-grams, etc. We report the result from their paper.
    
    \item \textbf{BiLSTM}~\cite{DBLP:conf/acl/LiuOSJ20}: The bidirectional LSTM has been widely used in sequence modeling as it explicitly stores historical information into memory and naturally captures context information from two directions.
    
    \item \textbf{HAN-BiLSTM}~\cite{DBLP:conf/acl/LiuOSJ20}: HAN~\cite{DBLP:conf/naacl/YangYDHSH16} computes document vectors in a hierarchical way of encoding and aggregation. This baseline replace its BiGRU with BiLSTM for the sake of comparison, and they also extend it with pretrained encoders and transformer layers.
    
    \item \textbf{Flat-MLMs.}~\cite{DBLP:conf/acl/LiuOSJ20}: Pretrained masked languages, e.g., RoBERTa, learn word representations and predict masked words by self-attention. This model use these encoders to encode the flat context concatenation like \text{[CTX]} $C^{0}$ \text{[SEP]} \text{[CTX]} $\cdots$ \text{[CTX]} $C^{l-1}$ \text{[SEP]} as Segment A and  \text{[CLS]} $C^l$ \text{[SEP]} as Segment B. After getting \text{[CTX]} and \text{[CLS]} representations, a gated transformer layer and a MLP predict impacts.
    As for XLNet, they follow its default setting so that \text{[CTX]} and \text{[CLS]} are located at the end of claims.
    
    \item \textbf{Interval-MLMs.}~\cite{DBLP:conf/acl/LiuOSJ20}: Flat-MLMs regard the context path as a whole segment and ignore the real discourse structures except the adjacency, e.g., distances between two claims are missing. They borrow the idea from BERT-SUM~\cite{DBLP:conf/emnlp/LiuL19}:
    segment embeddings of $C^{i}$ are assigned depending on whether the distance to $C^l$ is odd or even.
    \item \textbf{Context-MLMs.}~\cite{DBLP:conf/acl/LiuOSJ20}: This model is the pretrained encoders with context masks. A context mask is to localize the attention scope from the previous to the next. That is, $C^i$ can attends words in $C^{i-1}$ and $C^{i+1}$ except for itself if $1 \leq i < l$; $C^0$ can only attend $C^0, C^1$, and $C^l$ can only attend $C^{l-1}, C^{l}$.
    \item \textbf{Memory-MLMs.}~\cite{DBLP:conf/acl/LiuOSJ20}: XLNet utilizes memory to extend the capability of self-attention to learn super long historical text information.

    \item \textbf{BERT}~\cite{DBLP:conf/emnlp/DurmusLC19}: Fine-tuned a pre-trained deep bi-directional transformer language model (i.e., BERT)~\cite{DBLP:conf/naacl/DevlinCLT19} on the argument classification task, by adding a simple classification layer on top.
    
    \item \textbf{DISCOC}~\cite{DBLP:conf/acl/LiuOSJ20}: inject and fuse the sentence-level structural discourse information with contextualized features derived from large-scale language model (i.e.,Roberta).
    
    \item \textbf{Fine-Tuning(Flan-T5)}~\cite{DBLP:journals/jmlr/RaffelSRLNMZLL20}: Fine-tune a Flan-T5-model based on specifics tailored input text in various settings with a comparison of our model. The Implementation details are described in Appendix~\ref{sec: Appendix_t5_finetune_implementation}. 
    
    \item \textbf{Prefix-Tuning (Flan-T5)}~\cite{DBLP:conf/acl/LiL20}: a lightweight method concatenates the tunable prefix tokens before the discrete input text, keeps language model parameters frozen, and optimizes these continuous task-specific prefix tokens. The implementation details of the Prefix-Tuning methods are appended in Appendix~\ref{sec:implement_details_prompt_baseline}.
    
    \item \textbf{Prompt-Tuning (Flan-T5)}~\cite{DBLP:conf/emnlp/LesterAC21}: a vanilla Prompt Tuning-based model conditioning on a frozen model, releasing the constraints of the prompt templates from discrete to learnable prompts. The implementation details of the prompt tuning methods are appended in Appendix~\ref{sec:implement_details_prompt_baseline}.

\end{itemize}

\subsection{Implementation Details of the Prefix-Tuning and Prompt Tuning} \label{sec:implement_details_prompt_baseline}
In our paper, we implement the prefix tuning~\cite{DBLP:conf/acl/LiL20} and prompt tuning~\cite{DBLP:conf/emnlp/LesterAC21} methods as the baselines for comparison with our model. We proposed several templates to search for their best performance in these two methods. The experimental details for these two methods include the template and hyperparameter search. Moreover, there are 28 tokens, including textual tokens (non-tunable tokens) and tunable tokens, in our prompt template. For a fair comparison, we insert 28 tunable tokens into the respective prompt template in these two baselines.

\begin{figure*}[t]
    \centering
    \includegraphics[width=\linewidth]{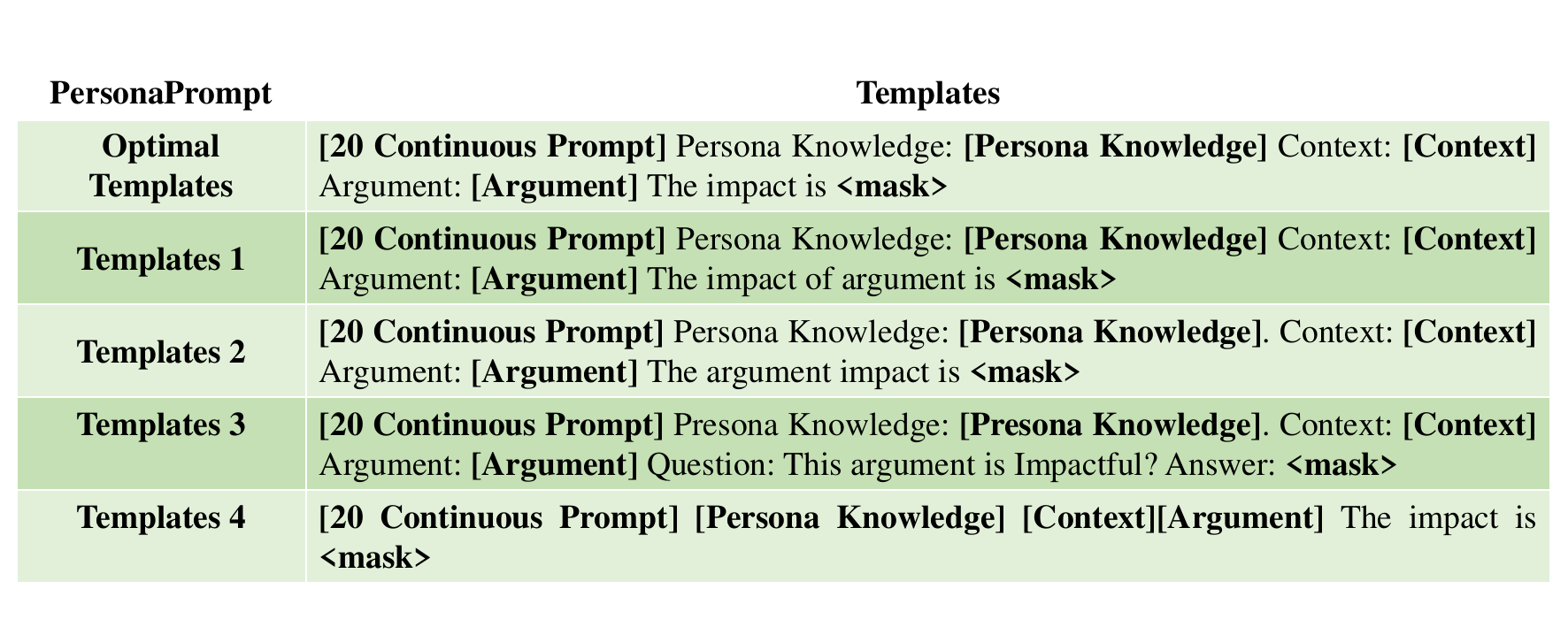}
    \vspace{-1.0 cm}
    \caption{PersonaPrompt Template Searching. The ``Optimal Templates'' is the finalized optimal template for implementing experiments to compare with extensive baselines.}
    \label{fig:PresonaPrompt_Template_Searching}
    \vspace{-0.5 cm}
\end{figure*}

\begin{table}[t]
\small
\centering
\scalebox{0.85}{
\begin{tabular}{l|c|c|c}
\toprule
\multicolumn{1}{c|}{Model} &  \multicolumn{1}{c|}{Precision} &  \multicolumn{1}{c|}{Recall} &  \multicolumn{1}{c}{F1} \\ 
\midrule
PersonaPrompt (Optimal) & \bf{64.35} & \bf{62.11} & \bf{62.95} \\
\midrule
PersonaPrompt (Template 1) & 61.18 & 62.00 & 62.13 \\
PersonaPrompt (Template 2) & 63.65 & 61.41 & 62.25 \\
PersonaPrompt (Template 3) & 63.55 & 59.67 & 61.61 \\
PersonaPrompt (Template 4) & 60.45 & 61.51 & 60.98 \\
\midrule
Continuous Prompt Length (10) & 62.48 & 61.55 & 61.73\\
Continuous Prompt Length (30) & 62.93 & 61.32 & 61.67\\
Continuous Prompt Length (50) & 63.72 & 61.48 & 62.32\\
\bottomrule
\end{tabular}
}
\vspace{-0.3cm}
\caption{Performance of prompt engineering on the PersonaPrompt (Flan-T5-base) in argument impact classification task. The upper part is prompt template searching on various templates, and the details of various templates are shown in Figure~\ref{fig:PresonaPrompt_Template_Searching}. The bottom part is the performance of various continuous prompt lengths in PersonaPrompt (Flan-T5-base) in the argument impact classification task. The default continuous prompt length of our model is 20.}
\label{table:Appendix_Prompt_Template_Searching}
\vspace{-0.1cm}
\end{table}

\begin{figure*}[t]
    \vspace{-0.4cm}
    \centering
    \includegraphics[width=\linewidth]{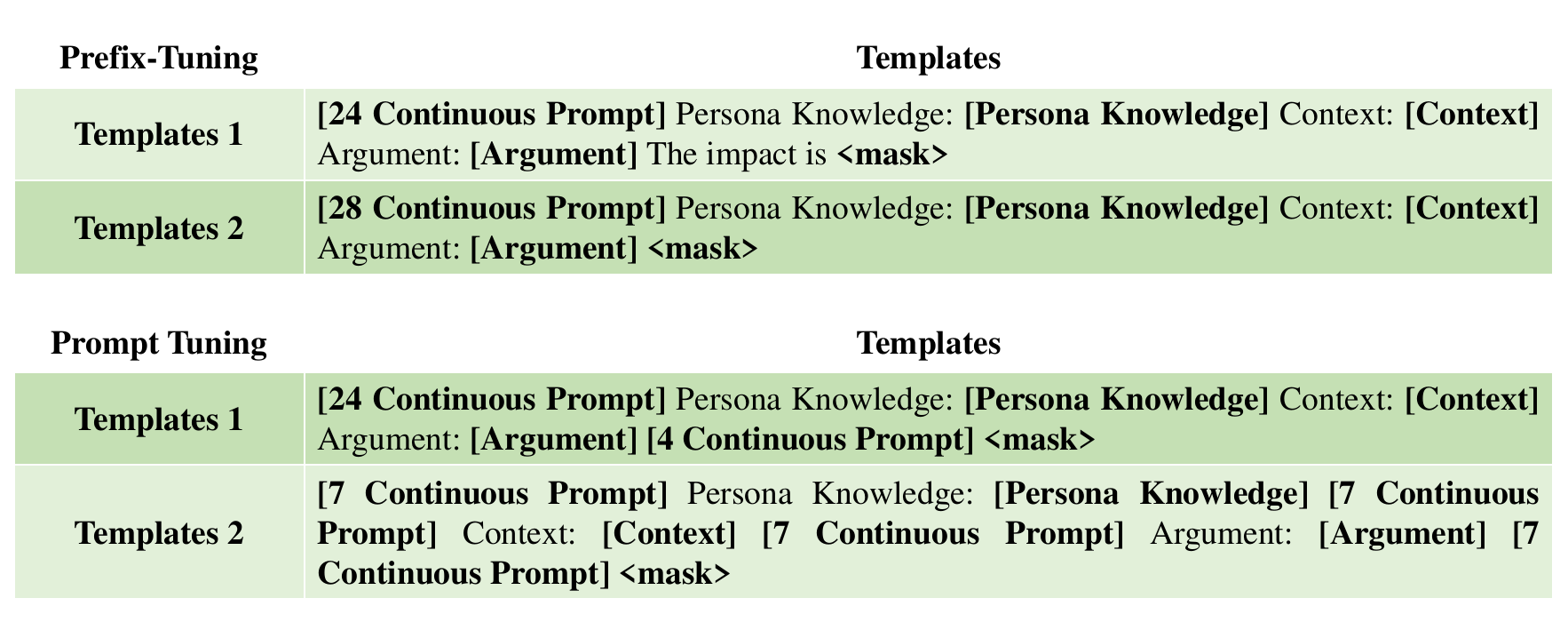}
    \vspace{-0.4cm}
    \caption{Prefix-Tuning and Prompt-Tuning Template Searching}
    \label{fig:Prefix_Prompt_Tuning_template}
    \vspace{-0.2 cm}
\end{figure*}

\paragraph{Prefix-Tuning} Following the setting of prefix tuning~\cite{DBLP:conf/acl/LiL20}, we implemented two designed templates on the argument impact classification task and the templates shown in figure~\ref{fig:Prefix_Prompt_Tuning_template}. In these templates, we find that the \textbf{prefix-prompt template one} is better among all templates, and we adopted this template for further comparison with our method. The overall configuration of this model follows the settings of prefix tuning~\cite{DBLP:conf/acl/LiL20}. The batch size and maximum sequence length of this model are 4 and 512. The training is performed using cross-entropy loss with an Adafactor optimizer~\cite{DBLP:conf/icml/ShazeerS18} and a learning rate selecting in {8e-5, 5e-5, 3e-5} yields the best performance on the validation set, and the training steps are 30,000. 

\paragraph{Prompt-Tuning} For the prompt tuning method, we implemented several designed templates on the argument impact classification task and the templates shown in figure~\ref{fig:Prefix_Prompt_Tuning_template}. In these templates, we find that the \textbf{prompt tuning template two} is better among all templates, and we adopted this template for further comparison with our method. The overall configuration of this model follows the settings of prompt tuning~\cite{DBLP:conf/emnlp/LesterAC21}. The batch size and maximum sequence length of this model are 4 and 512. The training is performed using cross-entropy loss with an Adafactor optimizer~\cite{DBLP:conf/icml/ShazeerS18} and a learning rate selecting in {8e-5, 5e-5, 3e-5} yields the best performance on the validation set, and the training steps are 30,000.

\subsection{Implementation Details of Flan-T5 Model Fine-Tuning} \label{sec: Appendix_t5_finetune_implementation}
Here we provide the fine-tuning details for Flan-T5-base and Flan-T5-large models on various datasets.

\noindent \textbf{Model Input and Output}
In main experiments, Flan-T5-model fine-tuning as the competitive baseline, we concatenate persona knowledge, context and arguments with an ``</s>'' at the end of the sequence as input. The Flan-T5 model asked to generate the class label (e.g., Impactful in argument classification task) given the data input.

\noindent \textbf{Hyperparameter Search}
We first conduct a preliminary experiment to determine the range of hyper-parameters. Then, we search for the learning rate within $\{3e-4, 1e-4\}$ and warmup steps within $\{0, 100\}$. For the Flan-T5-base model, we set the training batch size as 8, and the model is evaluated with a batch size of 128 every 150 steps. For the Flan-T5-large model, the training and evaluation batch sizes are set as 16 and 64, respectively. The model is optimized with an AdamW optimizer with a linear learning rate schedule. The test performance of the model with the best validation accuracy is reported.

\subsection{Implementation Details of Persona Knowledge with Large Language Models} \label{sec:Implementation_Details_of_LLMs}
\begin{figure}[!t]
    \centering
    \includegraphics[width=\linewidth]{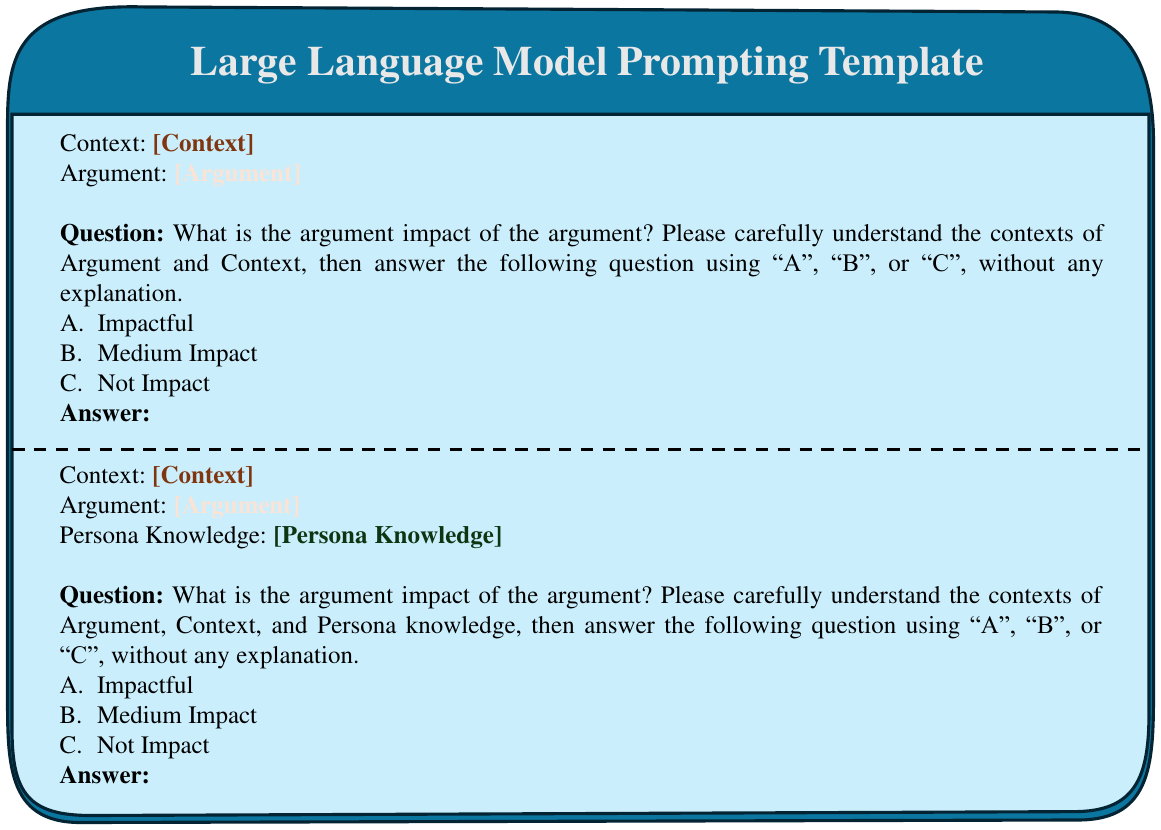}
    \vspace{-0.8 cm}
    \caption{Prompting template for large language models. The upper part is the baseline template refer to \citet{DBLP:conf/iclr/RobinsonW23}, and the bottom part is the prompt template with the persona knowledge.
    }
    \label{fig:ChatGPT_prompt_tempalte}
    \vspace{-0.6cm}
\end{figure}

While Large language models (LLMs) demonstrated remarkable proficiency across a diverse array of tasks~\cite{DBLP:journals/corr/abs-2303-12712,DBLP:journals/corr/abs-2302-04023,DBLP:conf/eacl/ChanCWJFLS24,DBLP:conf/acl/0001FLS0XWBLJCS24,DBLP:conf/ijcnlp/ChanLCCSWS23, DBLP:conf/coling/JiayangQC0SZ24}, certain obstacles persist unaddressed, including the inability to perform complex mathematical reasoning~\cite{DBLP:journals/corr/abs-2301-13867}, theory of mind reasoning~\cite{DBLP:journals/corr/abs-2404-13627,
lin2024constrainedreasoningchainsenhancing}, analogies reasoning~\cite{DBLP:conf/emnlp/ChengQCFWCRGZSZ23}, text-to-table generation~\cite{DBLP:journals/corr/abs-2404-14215}, associated ethical implications, and privacy preservation~\cite{DBLP:journals/corr/abs-2310-10383,DBLP:journals/corr/abs-2302-00539,DBLP:conf/acl/0003GLFH0CYYS24,DBLP:journals/corr/abs-2405-07667}. Therefore, we are interested in exploring the efficacy of large language models in zero-shot online debate tasks. The zero-shot performance of large language models, which relies on the sophisticated design of templates, has demonstrated variability across a range of tasks~\cite{DBLP:conf/acl/ChanLCLSWS23, DBLP:conf/naacl/MaZGTLZH22, DBLP:conf/icmlc2/ChanC23}.
To achieve replicable and representative results, we adopt the prompting template in \citet{DBLP:conf/iclr/RobinsonW23} to formulate the task as a multiple choice question answering problem as a baseline and append it with the audience persona knowledge to compare with this baseline. 
The prompting template is displayed in Figure~\ref{fig:ChatGPT_prompt_tempalte}.

\section{Appendix for Evaluation Result and Analysis}\label{sec:evaluation_result_analysis}

\begin{table*}[!t]
    \centering
    \small
    \vspace{-0.2cm}
    \scalebox{0.8}{\begin{tabular}{l|l|l|l}
    \toprule
    \multicolumn{1}{c|}{Model} &  \multicolumn{1}{c|}{Precision} &  \multicolumn{1}{c|}{Recall} &  \multicolumn{1}{c}{F1} \\ 
    \midrule
    \textsc{Majority} & 19.43 & 33.33 & 24.55 \\
    \textsc{SVM}~\cite{DBLP:conf/emnlp/DurmusLC19} & 65.67 & 38.58 & 35.42 \\
    \textsc{BiLSTM}~\cite{DBLP:conf/acl/LiuOSJ20} & 46.94 $\pm$ 1.08 & 46.64 $\pm$ 0.71 & 46.51 $\pm$ 1.11 \\
    \textsc{HAN-BiLSTM}~\cite{DBLP:conf/acl/LiuOSJ20} & 51.93 $\pm$ 1.37 & 49.08 $\pm$ 1.52 & 50.00 $\pm$ 1.49 \\
    \hline
    HAN-BERT~\cite{DBLP:conf/acl/LiuOSJ20} & 53.72 $\pm$ 0.80 & 53.45 $\pm$ 0.51 & 53.46 $\pm$ 0.47 \\
    HAN-RoBERTa~\cite{DBLP:conf/acl/LiuOSJ20} & 55.71 $\pm$ 1.12 & 55.95 $\pm$ 0.90 & 55.49 $\pm$ 0.62 \\
    HAN-XLNet~\cite{DBLP:conf/acl/LiuOSJ20}  & 53.91 $\pm$ 0.96 & 55.56 $\pm$ 1.59 & 54.53 $\pm$ 1.22 \\
    \hline
    \textsc{BERT~\cite{DBLP:conf/emnlp/DurmusLC19}} & 57.19 $\pm$ 0.92 & 55.77 $\pm$ 1.05 & 55.98 $\pm$ 0.70 \\
    \textsc{BERT (Knowledge)~\cite{DBLP:conf/emnlp/DurmusLC19}} & 56.76 $\pm$ 0.92 & 58.55 $\pm$ 1.16 & 57.25 $\pm$ 0.84 \\
    Flat-BERT~\cite{DBLP:conf/acl/LiuOSJ20}  & 57.34 $\pm$ 1.56 & 57.07 $\pm$ 0.74 & 56.75 $\pm$ 0.82 \\
    Flat-RoBERTa~\cite{DBLP:conf/acl/LiuOSJ20}  & 58.11 $\pm$ 1.34 & 56.40 $\pm$ 0.61 & 56.69 $\pm$ 0.63 \\
    Flat-XLNet~\cite{DBLP:conf/acl/LiuOSJ20} & 55.86 $\pm$ 1.74 & 56.20 $\pm$ 1.17 & 55.57 $\pm$ 0.95 \\
    \hline
    Interval-BERT~\cite{DBLP:conf/acl/LiuOSJ20} & 55.56 $\pm$ 2.03 & 55.52 $\pm$ 1.44 & 55.34 $\pm$ 1.50 \\
    Interval-RoBERTa~\cite{DBLP:conf/acl/LiuOSJ20} & 58.31 $\pm$ 0.89 & 56.46 $\pm$ 1.44 & 56.61 $\pm$ 1.24 \\
    Interval-XLNet~\cite{DBLP:conf/acl/LiuOSJ20} & 57.54 $\pm$ 0.50 & 56.78 $\pm$ 1.63 & 56.52 $\pm$ 1.00 \\
    \hline
    Context-BERT~\cite{DBLP:conf/acl/LiuOSJ20} & 54.96 $\pm$ 0.93 & 56.09 $\pm$ 0.83 & 55.44 $\pm$ 0.83 \\
    Context-RoBERTa~\cite{DBLP:conf/acl/LiuOSJ20} & 57.28 $\pm$ 0.97 & 55.29 $\pm$ 0.26 & 55.83 $\pm$ 0.54 \\
    Context-XLNet~\cite{DBLP:conf/acl/LiuOSJ20} & 54.56 $\pm$ 0.71 & 56.28 $\pm$ 1.22 & 55.10 $\pm$ 0.72 \\
    \hline
    Memory-BERT~\cite{DBLP:conf/acl/LiuOSJ20} & 54.33 $\pm$ 0.83 & 57.57 $\pm$ 0.67 & 55.22 $\pm$ 0.61 \\
    Memory-RoBERTa~\cite{DBLP:conf/acl/LiuOSJ20} & 55.08 $\pm$ 0.89 & 55.55 $\pm$ 1.59 & 54.76 $\pm$ 1.38 \\
    Memory-XLNet~\cite{DBLP:conf/acl/LiuOSJ20} & 55.44 $\pm$ 1.15 & 55.45 $\pm$ 1.25 & 54.91 $\pm$ 0.96 \\
    \hline
    Recurrent-BERT~\cite{DBLP:conf/acl/LiuOSJ20} & 56.37 $\pm$ 1.31 & 59.98 $\pm$ 1.29 & 57.45 $\pm$ 1.24 \\
    Recurrent-RoBERTa~\cite{DBLP:conf/acl/LiuOSJ20} & 59.28 $\pm$ 1.38 & 57.83 $\pm$ 0.84 & 57.81 $\pm$ 0.50 \\
    Recurrent-XLNet~\cite{DBLP:conf/acl/LiuOSJ20} & 56.09 $\pm$ 1.07 & 56.12 $\pm$ 1.35 & 55.68 $\pm$ 0.48 \\
    Recurrent-RoBERTa~\cite{DBLP:conf/acl/LiuOSJ20} & 59.94 $\pm$ 0.51 & 59.34 $\pm$ 1.28 & 58.52 $\pm$ 0.81 \\
    \textsc{DisCOC}~\cite{DBLP:conf/acl/LiuOSJ20} & 57.90 $\pm$ 0.70 & 59.41 $\pm$ 1.41 & 58.36 $\pm$ 0.52 \\
    \hline
    \textsc{Fine-Tuning (Flan-T5-base)}~\cite{DBLP:journals/jmlr/RaffelSRLNMZLL20} & 58.48 $\pm$ 1.04 & 59.57 $\pm$ 0.54 & 58.45 $\pm$ 0.69\\
    \textsc{Fine-Tuning (Flan-T5-base) (Knowledge)}~\cite{DBLP:journals/jmlr/RaffelSRLNMZLL20}  & 60.37 $\pm$ 1.73 & 60.58 $\pm$ 0.21 & 60.10 $\pm$ 1.25\\
    \textsc{Fine-Tuning (Flan-T5-large)~\cite{DBLP:journals/jmlr/RaffelSRLNMZLL20}} & 58.95 $\pm$ 0.81 & 63.42 $\pm$ 0.52 & 60.23 $\pm$ 0.23 \\
    \textsc{Fine-Tuning (Flan-T5-large) (Knowledge)~\cite{DBLP:journals/jmlr/RaffelSRLNMZLL20}} & 61.63 $\pm$ 1.04 & 66.56 $\pm$ 0.56 & 63.31 $\pm$ 0.68\\ 
    
    \textsc{Prefix-Tuning (Flan-T5-base)}~\cite{DBLP:conf/acl/LiL20} & 61.95 $\pm$ 2.03 & 57.69 $\pm$ 1.44 & 58.35 $\pm$ 1.24\\
    \textsc{Prefix-Tuning (Flan-T5-base) (Knowledge)}~\cite{DBLP:conf/acl/LiL20} & 61.85 $\pm$ 1.43 & 60.96 $\pm$ 0.17 & 60.13 $\pm$ 0.22\\
    \textsc{Prefix-Tuning (Flan-T5-large)}~\cite{DBLP:conf/acl/LiL20} & 65.78 $\pm$ 0.95 & 63.28 $\pm$ 1.25 & 63.75 $\pm$ 1.01\\
    \textsc{Prefix-Tuning (Flan-T5-large) (Knowledge)}~\cite{DBLP:conf/acl/LiL20} & 67.09 $\pm$ 1.53 & 65.07 $\pm$ 0.64 & 65.38 $\pm$ 0.71\\
    
    \textsc{Prompt-Tuning (Flan-T5-base)}~\cite{DBLP:conf/emnlp/LesterAC21} & 61.05 $\pm$ 1.58 & 57.80 $\pm$ 0.76 & 58.61 $\pm$ 0.84\\
    \textsc{Prompt-Tuning (Flan-T5-base) (Knowledge)}~\cite{DBLP:conf/emnlp/LesterAC21} & 60.52 $\pm$ 0.32 & 59.78 $\pm$ 0.62 & 59.58 $\pm$ 0.69\\
    \textsc{Prompt-Tuning (Flan-T5-large)}~\cite{DBLP:conf/emnlp/LesterAC21} & 63.48 $\pm$ 1.33 & 63.13 $\pm$ 0.78 & 63.10 $\pm$ 0.77\\  
    \textsc{Prompt-Tuning (Flan-T5-large) (Knowledge)}~\cite{DBLP:conf/emnlp/LesterAC21} & 66.79 $\pm$ 1.03 & 64.46 $\pm$ 0.75 & 65.28 $\pm$ 0.46\\
    \hline
    \textsc{PersonaPrompt(Flan-T5-base)} & 59.19 $\pm$ 0.57 & 60.20 $\pm$ 0.67 & 59.29 $\pm$ 0.67\\
    \textsc{PersonaPrompt(Flan-T5-base)(Knowledge)} & 64.35 $\pm$ 0.63 & 62.11 $\pm$ 0.36 & 62.81 $\pm$ 0.31\\
    \textsc{PersonaPrompt(Flan-T5-large)} & 65.40 $\pm$ 0.54 &64.26 $\pm$ 1.08 &  64.35 $\pm$ 0.51\\ 
    \textsc{PersonaPrompt(Flan-T5-large)(Knowledge)} & \bf{67.03} $\pm$ 1.24 & \bf{67.09} $\pm$ 0.58 &  \bf{67.67} $\pm$ 0.54 \\   
    \bottomrule
    \end{tabular}}
    \caption{The mean and standard deviation of different models' performance on the argument impact classification Task.}
    \label{table:arg_impact_supple}
\end{table*}

\subsection{Ablation study on the PresonaPrompt} \label{sec: Appendix_ablation_study_PersonaPrompt}

\paragraph{Prompt Template Searching}
We perform the prompt template research on our designed prompt template, and all prompt searching templates are enumerated in Figure~\ref{fig:PresonaPrompt_Template_Searching}, and the performance is shown in Table~\ref{table:Appendix_Prompt_Template_Searching}. Our finalized optimal template performs better than other templates, indicating the effectiveness of our tailored discrete tokens in the prompt template.

\paragraph{Continuous Prompt Length}
The continuous prompt (i.e., learnable prompt tokens) length is another factor that influences the performance of PersonaPrompt model. Hence, we implement various prompt lengths of 10, 20, 30, and 50. The performance is in Table~\ref{table:Appendix_Prompt_Template_Searching}, and the optimal continuous prompt length is 20, which provides the best performance among all the prompt lengths and is the default prompt length for implementing other experiments. Adopting a prompt length of more than 20 on PersonaPrompt will not significantly increase this task's performance on various evaluation metrics.

\subsection{More Case on the Attention Visualization}\label{sec:More_case_Attention_Visualization}
\begin{figure*}[t]
    \vspace{-0.3cm}
    \centering
    \includegraphics[width=\linewidth]{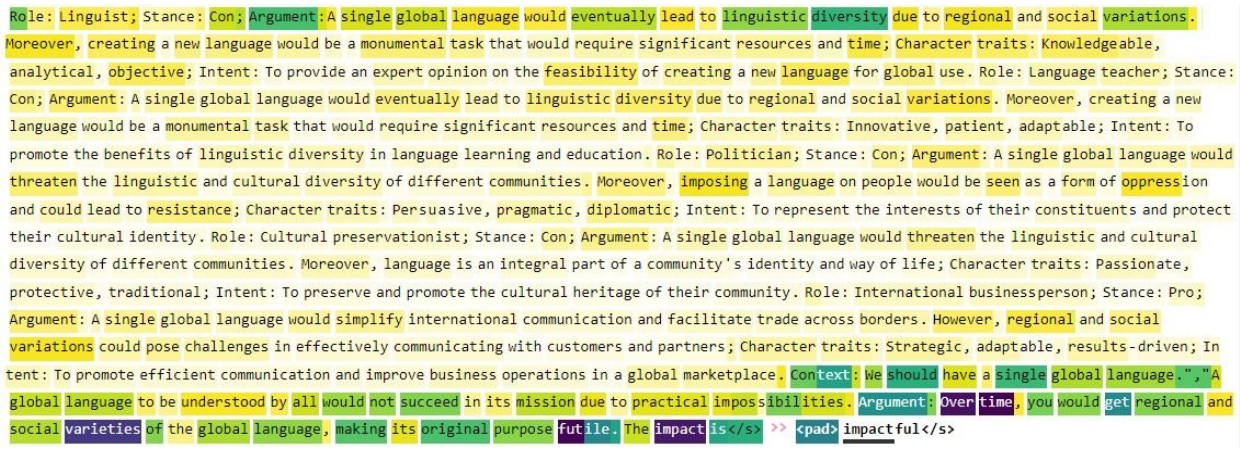}
    \vspace{-0.2cm}
    \caption{Attention Visualization for \textsc{Fine-Tuning (Flan-T5-base) (Knowledge)}.
    }\label{fig:Attention_Visualization_1}
    \vspace{-0.4cm}
\end{figure*}

\begin{figure*}[t]
    \centering
    \includegraphics[width=\linewidth]{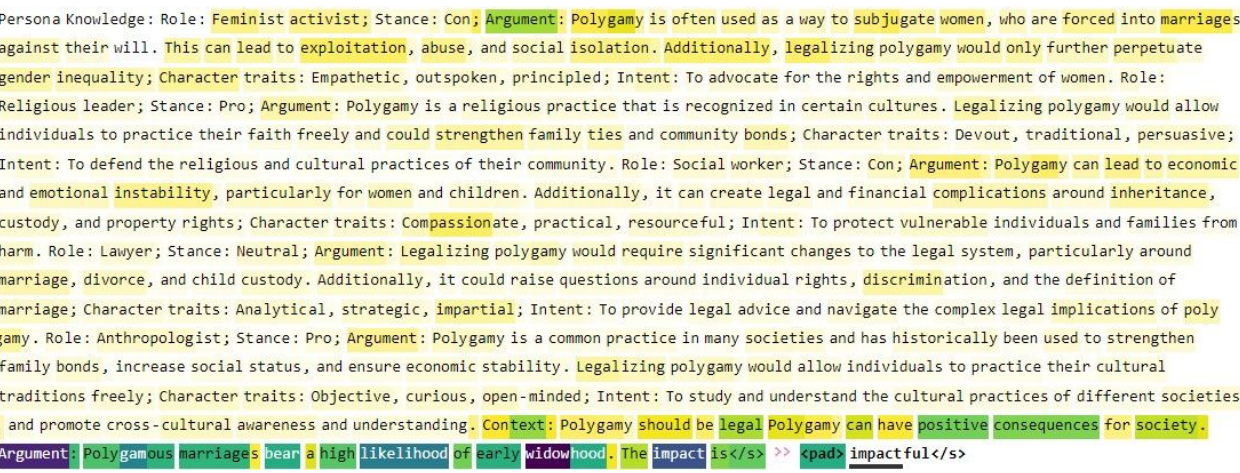}
    \vspace{-0.2cm}
    \caption{Attention Visualization for \textsc{Fine-Tuning (Flan-T5-base) (Knowledge)}.
    }\label{fig:Attention_Visualization_2}
    \vspace{-0.4cm}
\end{figure*}

More case studies on attention visualization are shown in Figure~\ref{fig:Attention_Visualization_1} and Figure~\ref{fig:Attention_Visualization_2}.

\subsection{More Case on the Retrieved ConceptNet Knowledge}\label{sec:More_case_ConceptNet}
\begin{figure*}[t]
    \centering
    \includegraphics[width=0.8\linewidth]{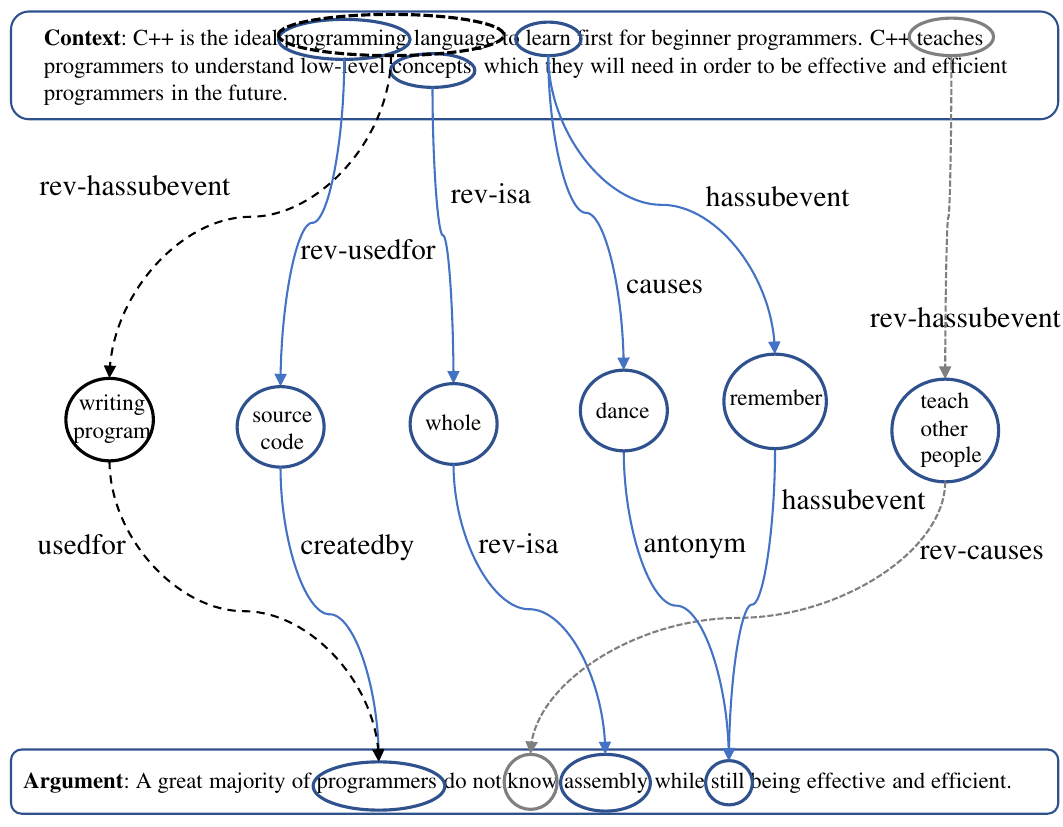}
    \vspace{-0.2cm}
    \caption{ConceptNet Knowledge grounded on the example shown in Figure 1 in the main context paper.}
    \label{fig:ConceptNet_Knowledge}
    \vspace{-0.4cm}
\end{figure*}

\begin{figure*}[t]
    \centering
    \includegraphics[width=0.8\linewidth]{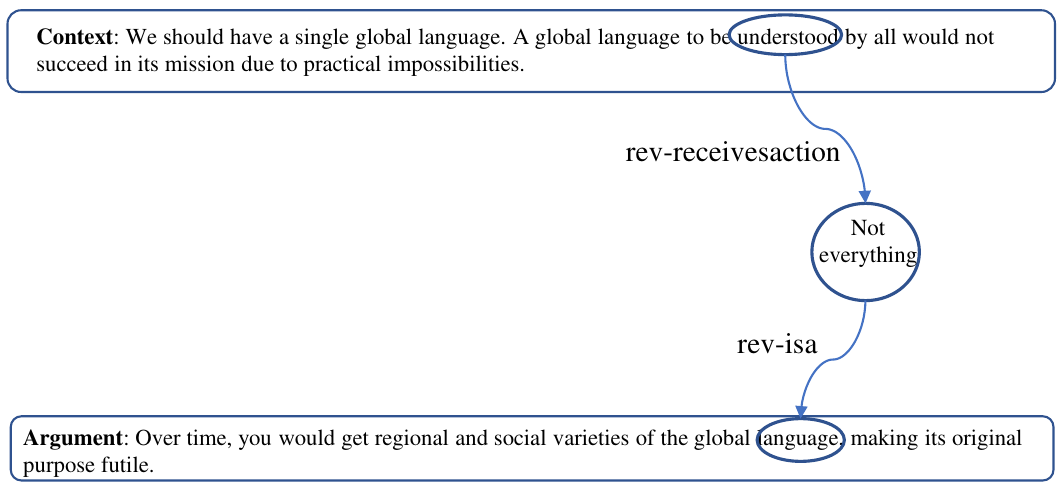}
    \vspace{-0.2cm}
    \caption{ConceptNet knowledge grounded on the example shown in Figure~\ref{fig:Attention_Visualization_2}.}
    \label{fig:ConceptNet_Knowledge_1}
    \vspace{-0.4cm}
\end{figure*}

More case studies on ConceptNet Knowledge grounded on data instance are shown in Figure~\ref{fig:ConceptNet_Knowledge} and Figure~\ref{fig:ConceptNet_Knowledge_1}.

%% file: emnlp2023.bbl
\begin{thebibliography}{80}
\expandafter\ifx\csname natexlab\endcsname\relax\def\natexlab#1{#1}\fi

\bibitem[{Alammar(2020)}]{alammar2020explaining}
J~Alammar. 2020.
\newblock \href {https://jalammar.github.io/explaining-transformers/} {Interfaces for explaining transformer language models}.

\bibitem[{Alshomary and Wachsmuth(2021)}]{DBLP:journals/patterns/AlshomaryW21}
Milad Alshomary and Henning Wachsmuth. 2021.
\newblock Toward audience-aware argument generation.
\newblock \emph{Patterns}, 2(6):100253.

\bibitem[{Baff et~al.(2020)Baff, Wachsmuth, Khatib, and Stein}]{DBLP:conf/acl/BaffWKS20}
Roxanne~El Baff, Henning Wachsmuth, Khalid~Al Khatib, and Benno Stein. 2020.
\newblock \href {https://doi.org/10.18653/v1/2020.acl-main.287} {Analyzing the persuasive effect of style in news editorial argumentation}.
\newblock In \emph{Proceedings of the 58th Annual Meeting of the Association for Computational Linguistics, {ACL} 2020, Online, July 5-10, 2020}, pages 3154--3160. Association for Computational Linguistics.

\bibitem[{Bang et~al.(2023)Bang, Cahyawijaya, Lee, Dai, Su, Wilie, Lovenia, Ji, Yu, Chung, Do, Xu, and Fung}]{DBLP:journals/corr/abs-2302-04023}
Yejin Bang, Samuel Cahyawijaya, Nayeon Lee, Wenliang Dai, Dan Su, Bryan Wilie, Holy Lovenia, Ziwei Ji, Tiezheng Yu, Willy Chung, Quyet~V. Do, Yan Xu, and Pascale Fung. 2023.
\newblock \href {http://arxiv.org/abs/2302.04023} {A multitask, multilingual, multimodal evaluation of chatgpt on reasoning, hallucination, and interactivity}.
\newblock \emph{CoRR}, abs/2302.04023.

\bibitem[{Brown et~al.(2020)Brown, Mann, Ryder, Subbiah, Kaplan, Dhariwal, Neelakantan, Shyam, Sastry, Askell, Agarwal, Herbert{-}Voss, Krueger, Henighan, Child, Ramesh, Ziegler, Wu, Winter, Hesse, Chen, Sigler, Litwin, Gray, Chess, Clark, Berner, McCandlish, Radford, Sutskever, and Amodei}]{brown2020language}
Tom~B. Brown, Benjamin Mann, Nick Ryder, Melanie Subbiah, Jared Kaplan, Prafulla Dhariwal, Arvind Neelakantan, Pranav Shyam, Girish Sastry, Amanda Askell, Sandhini Agarwal, Ariel Herbert{-}Voss, Gretchen Krueger, Tom Henighan, Rewon Child, Aditya Ramesh, Daniel~M. Ziegler, Jeffrey Wu, Clemens Winter, Christopher Hesse, Mark Chen, Eric Sigler, Mateusz Litwin, Scott Gray, Benjamin Chess, Jack Clark, Christopher Berner, Sam McCandlish, Alec Radford, Ilya Sutskever, and Dario Amodei. 2020.
\newblock Language models are few-shot learners.
\newblock In \emph{{NeurIPS}}.

\bibitem[{Bubeck et~al.(2023)Bubeck, Chandrasekaran, Eldan, Gehrke, Horvitz, Kamar, Lee, Lee, Li, Lundberg, Nori, Palangi, Ribeiro, and Zhang}]{DBLP:journals/corr/abs-2303-12712}
S{\'{e}}bastien Bubeck, Varun Chandrasekaran, Ronen Eldan, Johannes Gehrke, Eric Horvitz, Ece Kamar, Peter Lee, Yin~Tat Lee, Yuanzhi Li, Scott~M. Lundberg, Harsha Nori, Hamid Palangi, Marco~T{\'{u}}lio Ribeiro, and Yi~Zhang. 2023.
\newblock \href {http://arxiv.org/abs/2303.12712} {Sparks of artificial general intelligence: Early experiments with {GPT-4}}.
\newblock \emph{CoRR}, abs/2303.12712.

\bibitem[{Chan and Chan(2023)}]{DBLP:conf/icmlc2/ChanC23}
Chunkit Chan and Tsz~Ho Chan. 2023.
\newblock \href {https://doi.org/10.1145/3587716.3587743} {Discourse-aware prompt for argument impact classification}.
\newblock In \emph{Proceedings of the 15th International Conference on Machine Learning and Computing, {ICMLC} 2023, Zhuhai, China, February 17-20, 2023}, pages 165--171. {ACM}.

\bibitem[{Chan et~al.(2024{\natexlab{a}})Chan, Cheng, Wang, Jiang, Fang, Liu, and Song}]{DBLP:conf/eacl/ChanCWJFLS24}
Chunkit Chan, Jiayang Cheng, Weiqi Wang, Yuxin Jiang, Tianqing Fang, Xin Liu, and Yangqiu Song. 2024{\natexlab{a}}.
\newblock \href {https://aclanthology.org/2024.findings-eacl.47} {Exploring the potential of chatgpt on sentence level relations: {A} focus on temporal, causal, and discourse relations}.
\newblock In \emph{Findings of the Association for Computational Linguistics: {EACL} 2024, St. Julian's, Malta, March 17-22, 2024}, pages 684--721. Association for Computational Linguistics.

\bibitem[{Chan et~al.(2024{\natexlab{b}})Chan, Jiayang, Yim, Deng, Fan, Li, Liu, Zhang, Wang, and Song}]{DBLP:journals/corr/abs-2404-13627}
Chunkit Chan, Cheng Jiayang, Yauwai Yim, Zheye Deng, Wei Fan, Haoran Li, Xin Liu, Hongming Zhang, Weiqi Wang, and Yangqiu Song. 2024{\natexlab{b}}.
\newblock \href {https://doi.org/10.48550/ARXIV.2404.13627} {Negotiationtom: {A} benchmark for stress-testing machine theory of mind on negotiation surrounding}.
\newblock \emph{CoRR}, abs/2404.13627.

\bibitem[{Chan et~al.(2023{\natexlab{a}})Chan, Liu, Chan, Cheng, Song, Wong, and See}]{DBLP:conf/ijcnlp/ChanLCCSWS23}
Chunkit Chan, Xin Liu, Tsz~Ho Chan, Jiayang Cheng, Yangqiu Song, Ginny~Y. Wong, and Simon See. 2023{\natexlab{a}}.
\newblock \href {https://doi.org/10.18653/V1/2023.IJCNLP-MAIN.67} {Self-consistent narrative prompts on abductive natural language inference}.
\newblock In \emph{Proceedings of the 13th International Joint Conference on Natural Language Processing and the 3rd Conference of the Asia-Pacific Chapter of the Association for Computational Linguistics, {IJCNLP} 2023 -Volume 1: Long Papers, Nusa Dua, Bali, November 1 - 4, 2023}, pages 1040--1057. Association for Computational Linguistics.

\bibitem[{Chan et~al.(2023{\natexlab{b}})Chan, Liu, Cheng, Li, Song, Wong, and See}]{DBLP:conf/acl/ChanLCLSWS23}
Chunkit Chan, Xin Liu, Jiayang Cheng, Zihan Li, Yangqiu Song, Ginny~Y. Wong, and Simon See. 2023{\natexlab{b}}.
\newblock \href {https://doi.org/10.18653/v1/2023.findings-acl.4} {Discoprompt: Path prediction prompt tuning for implicit discourse relation recognition}.
\newblock In \emph{Findings of the Association for Computational Linguistics: {ACL} 2023, Toronto, Canada, July 9-14, 2023}, pages 35--57. Association for Computational Linguistics.

\bibitem[{Cheng et~al.(2023)Cheng, Qiu, Chan, Fang, Wang, Chan, Ru, Guo, Zhang, Song, Zhang, and Zhang}]{DBLP:conf/emnlp/ChengQCFWCRGZSZ23}
Jiayang Cheng, Lin Qiu, Tsz~Ho Chan, Tianqing Fang, Weiqi Wang, Chunkit Chan, Dongyu Ru, Qipeng Guo, Hongming Zhang, Yangqiu Song, Yue Zhang, and Zheng Zhang. 2023.
\newblock \href {https://doi.org/10.18653/V1/2023.EMNLP-MAIN.706} {Storyanalogy: Deriving story-level analogies from large language models to unlock analogical understanding}.
\newblock In \emph{Proceedings of the 2023 Conference on Empirical Methods in Natural Language Processing, {EMNLP} 2023, Singapore, December 6-10, 2023}, pages 11518--11537. Association for Computational Linguistics.

\bibitem[{Christiano et~al.(2017)Christiano, Leike, Brown, Martic, Legg, and Amodei}]{DBLP:conf/nips/ChristianoLBMLA17}
Paul~F. Christiano, Jan Leike, Tom~B. Brown, Miljan Martic, Shane Legg, and Dario Amodei. 2017.
\newblock Deep reinforcement learning from human preferences.
\newblock In \emph{Advances in Neural Information Processing Systems 30: Annual Conference on Neural Information Processing Systems 2017, December 4-9, 2017, Long Beach, CA, {USA}}, pages 4299--4307.

\bibitem[{Chung et~al.(2022)Chung, Hou, Longpre, Zoph, Tay, Fedus, Li, Wang, Dehghani, Brahma, Webson, Gu, Dai, Suzgun, Chen, Chowdhery, Narang, Mishra, Yu, Zhao, Huang, Dai, Yu, Petrov, Chi, Dean, Devlin, Roberts, Zhou, Le, and Wei}]{DBLP:journals/corr/abs-2210-11416}
Hyung~Won Chung, Le~Hou, Shayne Longpre, Barret Zoph, Yi~Tay, William Fedus, Eric Li, Xuezhi Wang, Mostafa Dehghani, Siddhartha Brahma, Albert Webson, Shixiang~Shane Gu, Zhuyun Dai, Mirac Suzgun, Xinyun Chen, Aakanksha Chowdhery, Sharan Narang, Gaurav Mishra, Adams Yu, Vincent~Y. Zhao, Yanping Huang, Andrew~M. Dai, Hongkun Yu, Slav Petrov, Ed~H. Chi, Jeff Dean, Jacob Devlin, Adam Roberts, Denny Zhou, Quoc~V. Le, and Jason Wei. 2022.
\newblock \href {https://doi.org/10.48550/arXiv.2210.11416} {Scaling instruction-finetuned language models}.
\newblock \emph{CoRR}, abs/2210.11416.

\bibitem[{Davison et~al.(2019)Davison, Feldman, and Rush}]{DBLP:conf/emnlp/DavisonFR19}
Joe Davison, Joshua Feldman, and Alexander~M. Rush. 2019.
\newblock Commonsense knowledge mining from pretrained models.
\newblock In \emph{Proceedings of the 2019 Conference on Empirical Methods in Natural Language Processing and the 9th International Joint Conference on Natural Language Processing, {EMNLP-IJCNLP} 2019}, pages 1173--1178.

\bibitem[{Deng et~al.(2024)Deng, Chan, Wang, Sun, Fan, Zheng, Yim, and Song}]{DBLP:journals/corr/abs-2404-14215}
Zheye Deng, Chunkit Chan, Weiqi Wang, Yuxi Sun, Wei Fan, Tianshi Zheng, Yauwai Yim, and Yangqiu Song. 2024.
\newblock \href {https://doi.org/10.48550/ARXIV.2404.14215} {Text-tuple-table: Towards information integration in text-to-table generation via global tuple extraction}.
\newblock \emph{CoRR}, abs/2404.14215.

\bibitem[{Devlin et~al.(2019)Devlin, Chang, Lee, and Toutanova}]{DBLP:conf/naacl/DevlinCLT19}
Jacob Devlin, Ming{-}Wei Chang, Kenton Lee, and Kristina Toutanova. 2019.
\newblock {BERT:} pre-training of deep bidirectional transformers for language understanding.
\newblock In \emph{Proceedings of the 2019 Conference of the North American Chapter of the Association for Computational Linguistics: Human Language Technologies, {NAACL-HLT} 2019, Minneapolis, MN, USA, June 2-7, 2019, Volume 1 (Long and Short Papers)}, pages 4171--4186.

\bibitem[{Ding et~al.(2022)Ding, Hu, Zhao, Chen, Liu, Zheng, and Sun}]{DBLP:conf/acl/DingHZCLZS22}
Ning Ding, Shengding Hu, Weilin Zhao, Yulin Chen, Zhiyuan Liu, Haitao Zheng, and Maosong Sun. 2022.
\newblock Openprompt: An open-source framework for prompt-learning.
\newblock In \emph{{ACL}}, pages 105--113.

\bibitem[{Du et~al.(2023)Du, Li, Torralba, Tenenbaum, and Mordatch}]{DBLP:journals/corr/abs-2305-14325}
Yilun Du, Shuang Li, Antonio Torralba, Joshua~B. Tenenbaum, and Igor Mordatch. 2023.
\newblock \href {http://arxiv.org/abs/2305.14325} {Improving factuality and reasoning in language models through multiagent debate}.
\newblock \emph{CoRR}, abs/2305.14325.

\bibitem[{Durmus and Cardie(2018)}]{DBLP:conf/naacl/DurmusC18}
Esin Durmus and Claire Cardie. 2018.
\newblock \href {https://doi.org/10.18653/v1/n18-1094} {Exploring the role of prior beliefs for argument persuasion}.
\newblock In \emph{Proceedings of the 2018 Conference of the North American Chapter of the Association for Computational Linguistics: Human Language Technologies, {NAACL-HLT} 2018, New Orleans, Louisiana, USA, June 1-6, 2018, Volume 1 (Long Papers)}, pages 1035--1045. Association for Computational Linguistics.

\bibitem[{Durmus and Cardie(2019{\natexlab{a}})}]{DBLP:conf/acl/DurmusC19}
Esin Durmus and Claire Cardie. 2019{\natexlab{a}}.
\newblock \href {https://doi.org/10.18653/v1/p19-1057} {A corpus for modeling user and language effects in argumentation on online debating}.
\newblock In \emph{Proceedings of the 57th Conference of the Association for Computational Linguistics, {ACL} 2019, Florence, Italy, July 28- August 2, 2019, Volume 1: Long Papers}, pages 602--607. Association for Computational Linguistics.

\bibitem[{Durmus and Cardie(2019{\natexlab{b}})}]{DBLP:conf/www/DurmusC19}
Esin Durmus and Claire Cardie. 2019{\natexlab{b}}.
\newblock \href {https://doi.org/10.1145/3308558.3313676} {Modeling the factors of user success in online debate}.
\newblock In \emph{The World Wide Web Conference, {WWW} 2019, San Francisco, CA, USA, May 13-17, 2019}, pages 2701--2707. {ACM}.

\bibitem[{Durmus et~al.(2019)Durmus, Ladhak, and Cardie}]{DBLP:conf/emnlp/DurmusLC19}
Esin Durmus, Faisal Ladhak, and Claire Cardie. 2019.
\newblock \href {https://doi.org/10.18653/v1/D19-1568} {The role of pragmatic and discourse context in determining argument impact}.
\newblock In \emph{Proceedings of the 2019 Conference on Empirical Methods in Natural Language Processing and the 9th International Joint Conference on Natural Language Processing, {EMNLP-IJCNLP} 2019, Hong Kong, China, November 3-7, 2019}, pages 5667--5677. Association for Computational Linguistics.

\bibitem[{Fleiss(1971)}]{fleiss1971measuring}
Joseph~L Fleiss. 1971.
\newblock Measuring nominal scale agreement among many raters.
\newblock \emph{Psychological bulletin}, 76(5):378.

\bibitem[{Frieder et~al.(2023)Frieder, Pinchetti, Griffiths, Salvatori, Lukasiewicz, Petersen, Chevalier, and Berner}]{DBLP:journals/corr/abs-2301-13867}
Simon Frieder, Luca Pinchetti, Ryan{-}Rhys Griffiths, Tommaso Salvatori, Thomas Lukasiewicz, Philipp~Christian Petersen, Alexis Chevalier, and Julius Berner. 2023.
\newblock \href {http://arxiv.org/abs/2301.13867} {Mathematical capabilities of chatgpt}.
\newblock \emph{CoRR}, abs/2301.13867.

\bibitem[{Gao et~al.(2023)Gao, Borges, Oh, Bayazit, Kanno, Wakaki, Mitsufuji, and Bosselut}]{DBLP:journals/corr/abs-2305-02364}
Silin Gao, Beatriz Borges, Soyoung Oh, Deniz Bayazit, Saya Kanno, Hiromi Wakaki, Yuki Mitsufuji, and Antoine Bosselut. 2023.
\newblock \href {http://arxiv.org/abs/2305.02364} {Peacok: Persona commonsense knowledge for consistent and engaging narratives}.
\newblock \emph{CoRR}, abs/2305.02364.

\bibitem[{Gordon and Durme(2013)}]{gordon2013reporting}
Jonathan Gordon and Benjamin~Van Durme. 2013.
\newblock Reporting bias and knowledge acquisition.
\newblock In \emph{Workshop on AKBC@CIKM}, pages 25--30.

\bibitem[{Habernal and Gurevych(2016{\natexlab{a}})}]{DBLP:conf/emnlp/HabernalG16}
Ivan Habernal and Iryna Gurevych. 2016{\natexlab{a}}.
\newblock What makes a convincing argument? empirical analysis and detecting attributes of convincingness in web argumentation.
\newblock In \emph{Proceedings of the 2016 Conference on Empirical Methods in Natural Language Processing, {EMNLP} 2016, Austin, Texas, USA, November 1-4, 2016}, pages 1214--1223.

\bibitem[{Habernal and Gurevych(2016{\natexlab{b}})}]{DBLP:conf/acl/HabernalG16}
Ivan Habernal and Iryna Gurevych. 2016{\natexlab{b}}.
\newblock Which argument is more convincing? analyzing and predicting convincingness of web arguments using bidirectional {LSTM}.
\newblock In \emph{Proceedings of the 54th Annual Meeting of the Association for Computational Linguistics, {ACL} 2016, August 7-12, 2016, Berlin, Germany, Volume 1: Long Papers}.

\bibitem[{Hidey and McKeown(2018)}]{DBLP:conf/aaai/HideyM18}
Christopher Hidey and Kathleen~R. McKeown. 2018.
\newblock Persuasive influence detection: The role of argument sequencing.
\newblock In \emph{Proceedings of the Thirty-Second {AAAI} Conference on Artificial Intelligence, (AAAI-18), the 30th innovative Applications of Artificial Intelligence (IAAI-18), and the 8th {AAAI} Symposium on Educational Advances in Artificial Intelligence (EAAI-18)}, pages 5173--5180.

\bibitem[{Jiang et~al.(2023{\natexlab{a}})Jiang, Zhang, Cao, Kabbara, and Roy}]{jiang2023personallm}
Hang Jiang, Xiajie Zhang, Xubo Cao, Jad Kabbara, and Deb Roy. 2023{\natexlab{a}}.
\newblock Personallm: Investigating the ability of gpt-3.5 to express personality traits and gender differences.
\newblock \emph{arXiv preprint arXiv:2305.02547}.

\bibitem[{Jiang et~al.(2023{\natexlab{b}})Jiang, Chan, Chen, and Wang}]{DBLP:conf/emnlp/JiangCCW23}
Yuxin Jiang, Chunkit Chan, Mingyang Chen, and Wei Wang. 2023{\natexlab{b}}.
\newblock \href {https://doi.org/10.18653/V1/2023.EMNLP-MAIN.189} {Lion: Adversarial distillation of proprietary large language models}.
\newblock In \emph{Proceedings of the 2023 Conference on Empirical Methods in Natural Language Processing, {EMNLP} 2023, Singapore, December 6-10, 2023}, pages 3134--3154. Association for Computational Linguistics.

\bibitem[{Jiang et~al.(2020)Jiang, Xu, Araki, and Neubig}]{DBLP:journals/tacl/JiangXAN20}
Zhengbao Jiang, Frank~F. Xu, Jun Araki, and Graham Neubig. 2020.
\newblock How can we know what language models know.
\newblock \emph{Trans. Assoc. Comput. Linguistics}, 8:423--438.

\bibitem[{Jiayang et~al.(2024)Jiayang, Qiu, Chan, Liu, Song, and Zhang}]{DBLP:conf/coling/JiayangQC0SZ24}
Cheng Jiayang, Lin Qiu, Chunkit Chan, Xin Liu, Yangqiu Song, and Zheng Zhang. 2024.
\newblock \href {https://aclanthology.org/2024.lrec-main.587} {Eventground: Narrative reasoning by grounding to eventuality-centric knowledge graphs}.
\newblock In \emph{Proceedings of the 2024 Joint International Conference on Computational Linguistics, Language Resources and Evaluation, {LREC/COLING} 2024, 20-25 May, 2024, Torino, Italy}, pages 6622--6642. {ELRA} and {ICCL}.

\bibitem[{Johnson(2013)}]{johnson2013role}
Ralph~H Johnson. 2013.
\newblock The role of audience in argumentation from the perspective of informal logic.
\newblock \emph{Philosophy \& rhetoric}, 46(4):533--549.

\bibitem[{Lauscher et~al.(2022)Lauscher, Wachsmuth, Gurevych, and Glavas}]{DBLP:journals/tacl/LauscherWGG22}
Anne Lauscher, Henning Wachsmuth, Iryna Gurevych, and Goran Glavas. 2022.
\newblock \href {https://transacl.org/ojs/index.php/tacl/article/view/3967} {Scientia potentia est - on the role of knowledge in computational argumentation}.
\newblock \emph{Trans. Assoc. Comput. Linguistics}, 10:1392--1422.

\bibitem[{Lester et~al.(2021)Lester, Al{-}Rfou, and Constant}]{DBLP:conf/emnlp/LesterAC21}
Brian Lester, Rami Al{-}Rfou, and Noah Constant. 2021.
\newblock The power of scale for parameter-efficient prompt tuning.
\newblock In \emph{Proceedings of the 2021 Conference on Empirical Methods in Natural Language Processing, {EMNLP} 2021}, pages 3045--3059.

\bibitem[{Li et~al.(2023{\natexlab{a}})Li, Chen, Luo, Kang, Zhang, Hu, Chan, and Song}]{DBLP:journals/corr/abs-2310-10383}
Haoran Li, Yulin Chen, Jinglong Luo, Yan Kang, Xiaojin Zhang, Qi~Hu, Chunkit Chan, and Yangqiu Song. 2023{\natexlab{a}}.
\newblock \href {https://doi.org/10.48550/ARXIV.2310.10383} {Privacy in large language models: Attacks, defenses and future directions}.
\newblock \emph{CoRR}, abs/2310.10383.

\bibitem[{Li et~al.(2024{\natexlab{a}})Li, Chen, Zheng, Hu, Chan, Liu, and Song}]{DBLP:journals/corr/abs-2405-07667}
Haoran Li, Yulin Chen, Zihao Zheng, Qi~Hu, Chunkit Chan, Heshan Liu, and Yangqiu Song. 2024{\natexlab{a}}.
\newblock \href {https://doi.org/10.48550/ARXIV.2405.07667} {Backdoor removal for generative large language models}.
\newblock \emph{CoRR}, abs/2405.07667.

\bibitem[{Li et~al.(2024{\natexlab{b}})Li, Guo, Li, Fan, Hu, Liu, Chan, Yao, Yao, and Song}]{DBLP:conf/acl/0003GLFH0CYYS24}
Haoran Li, Dadi Guo, Donghao Li, Wei Fan, Qi~Hu, Xin Liu, Chunkit Chan, Duanyi Yao, Yuan Yao, and Yangqiu Song. 2024{\natexlab{b}}.
\newblock \href {https://aclanthology.org/2024.acl-long.4} {Privlm-bench: {A} multi-level privacy evaluation benchmark for language models}.
\newblock In \emph{Proceedings of the 62nd Annual Meeting of the Association for Computational Linguistics (Volume 1: Long Papers), {ACL} 2024, Bangkok, Thailand, August 11-16, 2024}, pages 54--73. Association for Computational Linguistics.

\bibitem[{Li et~al.(2020)Li, Durmus, and Cardie}]{DBLP:conf/emnlp/LiDC20}
Jialu Li, Esin Durmus, and Claire Cardie. 2020.
\newblock \href {https://doi.org/10.18653/v1/2020.emnlp-main.716} {Exploring the role of argument structure in online debate persuasion}.
\newblock In \emph{Proceedings of the 2020 Conference on Empirical Methods in Natural Language Processing, {EMNLP} 2020, Online, November 16-20, 2020}, pages 8905--8912. Association for Computational Linguistics.

\bibitem[{Li et~al.(2023{\natexlab{b}})Li, Cheng, Zhao, Nie, and Wen}]{DBLP:journals/corr/abs-2305-11747}
Junyi Li, Xiaoxue Cheng, Wayne~Xin Zhao, Jian{-}Yun Nie, and Ji{-}Rong Wen. 2023{\natexlab{b}}.
\newblock \href {https://doi.org/10.48550/arXiv.2305.11747} {Halueval: {A} large-scale hallucination evaluation benchmark for large language models}.
\newblock \emph{CoRR}, abs/2305.11747.

\bibitem[{Li and Liang(2021)}]{DBLP:conf/acl/LiL20}
Xiang~Lisa Li and Percy Liang. 2021.
\newblock \href {https://doi.org/10.18653/v1/2021.acl-long.353} {Prefix-tuning: Optimizing continuous prompts for generation}.
\newblock In \emph{Proceedings of the 59th Annual Meeting of the Association for Computational Linguistics and the 11th International Joint Conference on Natural Language Processing, {ACL/IJCNLP} 2021, (Volume 1: Long Papers), Virtual Event, August 1-6, 2021}, pages 4582--4597. Association for Computational Linguistics.

\bibitem[{Li et~al.(2024{\natexlab{c}})Li, Wen, Wang, Li, Yuan, Liu, Liu, Xu, Wang, Sun et~al.}]{li2024personal}
Yuanchun Li, Hao Wen, Weijun Wang, Xiangyu Li, Yizhen Yuan, Guohong Liu, Jiacheng Liu, Wenxing Xu, Xiang Wang, Yi~Sun, et~al. 2024{\natexlab{c}}.
\newblock Personal llm agents: Insights and survey about the capability, efficiency and security.
\newblock \emph{arXiv preprint arXiv:2401.05459}.

\bibitem[{Lin et~al.(2019)Lin, Chen, Chen, and Ren}]{DBLP:conf/emnlp/LinCCR19}
Bill~Yuchen Lin, Xinyue Chen, Jamin Chen, and Xiang Ren. 2019.
\newblock \href {https://doi.org/10.18653/v1/D19-1282} {Kagnet: Knowledge-aware graph networks for commonsense reasoning}.
\newblock In \emph{Proceedings of the 2019 Conference on Empirical Methods in Natural Language Processing and the 9th International Joint Conference on Natural Language Processing, {EMNLP-IJCNLP} 2019, Hong Kong, China, November 3-7, 2019}, pages 2829--2839. Association for Computational Linguistics.

\bibitem[{Lin et~al.(2024)Lin, Chan, Song, and Liu}]{lin2024constrainedreasoningchainsenhancing}
Zizheng Lin, Chunkit Chan, Yangqiu Song, and Xin Liu. 2024.
\newblock \href {http://arxiv.org/abs/2409.13490} {Constrained reasoning chains for enhancing theory-of-mind in large language models}.

\bibitem[{Liu et~al.(2022)Liu, Liu, Lu, Welleck, West, Bras, Choi, and Hajishirzi}]{DBLP:conf/acl/0010LLWWBCH22}
Jiacheng Liu, Alisa Liu, Ximing Lu, Sean Welleck, Peter West, Ronan~Le Bras, Yejin Choi, and Hannaneh Hajishirzi. 2022.
\newblock Generated knowledge prompting for commonsense reasoning.
\newblock In \emph{Proceedings of the 60th Annual Meeting of the Association for Computational Linguistics, {ACL} 2022}, pages 3154--3169.

\bibitem[{Liu et~al.(2021{\natexlab{a}})Liu, Yuan, Fu, Jiang, Hayashi, and Neubig}]{DBLP:journals/corr/abs-2107-13586}
Pengfei Liu, Weizhe Yuan, Jinlan Fu, Zhengbao Jiang, Hiroaki Hayashi, and Graham Neubig. 2021{\natexlab{a}}.
\newblock \href {http://arxiv.org/abs/2107.13586} {Pre-train, prompt, and predict: {A} systematic survey of prompting methods in natural language processing}.
\newblock \emph{CoRR}, abs/2107.13586.

\bibitem[{Liu et~al.(2021{\natexlab{b}})Liu, Ou, Song, and Jiang}]{DBLP:conf/acl/LiuOSJ20}
Xin Liu, Jiefu Ou, Yangqiu Song, and Xin Jiang. 2021{\natexlab{b}}.
\newblock \href {https://doi.org/10.18653/v1/2021.acl-long.306} {Exploring discourse structures for argument impact classification}.
\newblock In \emph{Proceedings of the 59th Annual Meeting of the Association for Computational Linguistics and the 11th International Joint Conference on Natural Language Processing, {ACL/IJCNLP} 2021, (Volume 1: Long Papers), Virtual Event, August 1-6, 2021}, pages 3958--3969. Association for Computational Linguistics.

\bibitem[{Liu and Lapata(2019)}]{DBLP:conf/emnlp/LiuL19}
Yang Liu and Mirella Lapata. 2019.
\newblock Text summarization with pretrained encoders.
\newblock In \emph{Proceedings of the 2019 Conference on Empirical Methods in Natural Language Processing and the 9th International Joint Conference on Natural Language Processing, {EMNLP-IJCNLP} 2019, Hong Kong, China, November 3-7, 2019}, pages 3728--3738.

\bibitem[{Lukas et~al.(2023)Lukas, Salem, Sim, Tople, Wutschitz, and B{\'{e}}guelin}]{DBLP:journals/corr/abs-2302-00539}
Nils Lukas, Ahmed Salem, Robert Sim, Shruti Tople, Lukas Wutschitz, and Santiago~Zanella B{\'{e}}guelin. 2023.
\newblock \href {http://arxiv.org/abs/2302.00539} {Analyzing leakage of personally identifiable information in language models}.
\newblock \emph{CoRR}, abs/2302.00539.

\bibitem[{Lukin et~al.(2017)Lukin, Anand, Walker, and Whittaker}]{DBLP:conf/eacl/WalkerALW17}
Stephanie~M. Lukin, Pranav Anand, Marilyn~A. Walker, and Steve Whittaker. 2017.
\newblock Argument strength is in the eye of the beholder: Audience effects in persuasion.
\newblock In \emph{Proceedings of the 15th Conference of the European Chapter of the Association for Computational Linguistics, {EACL} 2017, Valencia, Spain, April 3-7, 2017, Volume 1: Long Papers}, pages 742--753.

\bibitem[{Ma et~al.(2022)Ma, Zhou, Gui, Tan, Li, Zhang, and Huang}]{DBLP:conf/naacl/MaZGTLZH22}
Ruotian Ma, Xin Zhou, Tao Gui, Yiding Tan, Linyang Li, Qi~Zhang, and Xuanjing Huang. 2022.
\newblock \href {https://doi.org/10.18653/v1/2022.naacl-main.420} {Template-free prompt tuning for few-shot {NER}}.
\newblock In \emph{Proceedings of the 2022 Conference of the North American Chapter of the Association for Computational Linguistics: Human Language Technologies, {NAACL} 2022, Seattle, WA, United States, July 10-15, 2022}, pages 5721--5732. Association for Computational Linguistics.

\bibitem[{Moens(2018)}]{DBLP:journals/argcom/Moens18}
Marie{-}Francine Moens. 2018.
\newblock Argumentation mining: How can a machine acquire common sense and world knowledge?
\newblock \emph{Argument Comput.}, 9(1):1--14.

\bibitem[{Moore et~al.(2019)Moore, Barbour, and Marshall}]{moore2019persona}
Christopher Moore, Kim Barbour, and P~David Marshall. 2019.
\newblock \emph{Persona studies: an introduction}.
\newblock John Wiley \& Sons.

\bibitem[{OpenAI(2023)}]{DBLP:journals/corr/abs-2303-08774}
OpenAI. 2023.
\newblock \href {http://arxiv.org/abs/2303.08774} {{GPT-4} technical report}.
\newblock \emph{CoRR}, abs/2303.08774.

\bibitem[{OpenAI(2022)}]{openai2022chatgpt}
TB~OpenAI. 2022.
\newblock Chatgpt: Optimizing language models for dialogue.
\newblock \emph{OpenAI}.

\bibitem[{Paranjape et~al.(2021)Paranjape, Michael, Ghazvininejad, Hajishirzi, and Zettlemoyer}]{DBLP:conf/acl/ParanjapeMGHZ21}
Bhargavi Paranjape, Julian Michael, Marjan Ghazvininejad, Hannaneh Hajishirzi, and Luke Zettlemoyer. 2021.
\newblock \href {https://doi.org/10.18653/v1/2021.findings-acl.366} {Prompting contrastive explanations for commonsense reasoning tasks}.
\newblock In \emph{Findings of the Association for Computational Linguistics: {ACL/IJCNLP} 2021, Online Event, August 1-6, 2021}, volume {ACL/IJCNLP} 2021 of \emph{Findings of {ACL}}, pages 4179--4192. Association for Computational Linguistics.

\bibitem[{Park et~al.(2023)Park, O'Brien, Cai, Morris, Liang, and Bernstein}]{DBLP:journals/corr/abs-2304-03442}
Joon~Sung Park, Joseph~C. O'Brien, Carrie~J. Cai, Meredith~Ringel Morris, Percy Liang, and Michael~S. Bernstein. 2023.
\newblock \href {https://doi.org/10.48550/arXiv.2304.03442} {Generative agents: Interactive simulacra of human behavior}.
\newblock \emph{CoRR}, abs/2304.03442.

\bibitem[{Perez et~al.(2021)Perez, Kiela, and Cho}]{DBLP:conf/nips/PerezKC21}
Ethan Perez, Douwe Kiela, and Kyunghyun Cho. 2021.
\newblock True few-shot learning with language models.
\newblock In \emph{Advances in Neural Information Processing Systems 34: Annual Conference on Neural Information Processing Systems 2021, NeurIPS 2021}, pages 11054--11070.

\bibitem[{Petroni et~al.(2019)Petroni, Rockt{\"{a}}schel, Riedel, Lewis, Bakhtin, Wu, and Miller}]{DBLP:conf/emnlp/PetroniRRLBWM19}
Fabio Petroni, Tim Rockt{\"{a}}schel, Sebastian Riedel, Patrick S.~H. Lewis, Anton Bakhtin, Yuxiang Wu, and Alexander~H. Miller. 2019.
\newblock Language models as knowledge bases?
\newblock In \emph{Proceedings of the 2019 Conference on Empirical Methods in Natural Language Processing and the 9th International Joint Conference on Natural Language Processing, {EMNLP-IJCNLP} 2019}, pages 2463--2473.

\bibitem[{Petty and Cacioppo(1986)}]{petty1986elaboration}
Richard~E Petty and John~T Cacioppo. 1986.
\newblock The elaboration likelihood model of persuasion.
\newblock In \emph{Advances in Experimental Social Psychology}, volume~19, pages 123--205. Elsevier.

\bibitem[{Raffel et~al.(2020)Raffel, Shazeer, Roberts, Lee, Narang, Matena, Zhou, Li, and Liu}]{DBLP:journals/jmlr/RaffelSRLNMZLL20}
Colin Raffel, Noam Shazeer, Adam Roberts, Katherine Lee, Sharan Narang, Michael Matena, Yanqi Zhou, Wei Li, and Peter~J. Liu. 2020.
\newblock Exploring the limits of transfer learning with a unified text-to-text transformer.
\newblock \emph{J. Mach. Learn. Res.}, 21:140:1--140:67.

\bibitem[{Robinson and Wingate(2023)}]{DBLP:conf/iclr/RobinsonW23}
Joshua Robinson and David Wingate. 2023.
\newblock Leveraging large language models for multiple choice question answering.
\newblock In \emph{The Eleventh International Conference on Learning Representations, {ICLR} 2023}.

\bibitem[{Sap et~al.(2022)Sap, Swayamdipta, Vianna, Zhou, Choi, and Smith}]{DBLP:conf/naacl/SapSVZCS22}
Maarten Sap, Swabha Swayamdipta, Laura Vianna, Xuhui Zhou, Yejin Choi, and Noah~A. Smith. 2022.
\newblock \href {https://doi.org/10.18653/v1/2022.naacl-main.431} {Annotators with attitudes: How annotator beliefs and identities bias toxic language detection}.
\newblock In \emph{Proceedings of the 2022 Conference of the North American Chapter of the Association for Computational Linguistics: Human Language Technologies, {NAACL} 2022, Seattle, WA, United States, July 10-15, 2022}, pages 5884--5906. Association for Computational Linguistics.

\bibitem[{Schick et~al.(2021)Schick, Udupa, and Sch{\"{u}}tze}]{DBLP:journals/tacl/SchickUS21}
Timo Schick, Sahana Udupa, and Hinrich Sch{\"{u}}tze. 2021.
\newblock \href {https://doi.org/10.1162/tacl\_a\_00434} {Self-diagnosis and self-debiasing: {A} proposal for reducing corpus-based bias in {NLP}}.
\newblock \emph{Trans. Assoc. Comput. Linguistics}, 9:1408--1424.

\bibitem[{Shazeer and Stern(2018)}]{DBLP:conf/icml/ShazeerS18}
Noam Shazeer and Mitchell Stern. 2018.
\newblock Adafactor: Adaptive learning rates with sublinear memory cost.
\newblock In \emph{{ICML}}, pages 4603--4611.

\bibitem[{Shmueli{-}Scheuer et~al.(2019)Shmueli{-}Scheuer, Herzig, Konopnicki, and Sandbank}]{DBLP:conf/um/Shmueli-Scheuer19}
Michal Shmueli{-}Scheuer, Jonathan Herzig, David Konopnicki, and Tommy Sandbank. 2019.
\newblock \href {https://doi.org/10.1145/3320435.3320467} {Detecting persuasive arguments based on author-reader personality traits and their interaction}.
\newblock In \emph{Proceedings of the 27th {ACM} Conference on User Modeling, Adaptation and Personalization, {UMAP} 2019, Larnaca, Cyprus, June 9-12, 2019}, pages 211--215. {ACM}.

\bibitem[{Simpson and Gurevych(2018)}]{DBLP:journals/tacl/SimpsonG18}
Edwin~D. Simpson and Iryna Gurevych. 2018.
\newblock Finding convincing arguments using scalable bayesian preference learning.
\newblock \emph{Trans. Assoc. Comput. Linguistics}, 6:357--371.

\bibitem[{Speer et~al.(2017)Speer, Chin, and Havasi}]{DBLP:conf/aaai/SpeerCH17}
Robyn Speer, Joshua Chin, and Catherine Havasi. 2017.
\newblock \href {http://aaai.org/ocs/index.php/AAAI/AAAI17/paper/view/14972} {Conceptnet 5.5: An open multilingual graph of general knowledge}.
\newblock In \emph{Proceedings of the Thirty-First {AAAI} Conference on Artificial Intelligence, February 4-9, 2017, San Francisco, California, {USA}}, pages 4444--4451. {AAAI} Press.

\bibitem[{Tan et~al.(2016)Tan, Niculae, Danescu{-}Niculescu{-}Mizil, and Lee}]{DBLP:conf/www/TanNDL16}
Chenhao Tan, Vlad Niculae, Cristian Danescu{-}Niculescu{-}Mizil, and Lillian Lee. 2016.
\newblock \href {https://doi.org/10.1145/2872427.2883081} {Winning arguments: Interaction dynamics and persuasion strategies in good-faith online discussions}.
\newblock In \emph{Proceedings of the 25th International Conference on World Wide Web, {WWW} 2016, Montreal, Canada, April 11 - 15, 2016}, pages 613--624. {ACM}.

\bibitem[{Wang et~al.(2022)Wang, Huang, Qiu, Shi, Wang, Li, and Gao}]{DBLP:conf/emnlp/WangHQSWLG22}
Jianing Wang, Wenkang Huang, Minghui Qiu, Qiuhui Shi, Hongbin Wang, Xiang Li, and Ming Gao. 2022.
\newblock \href {https://aclanthology.org/2022.emnlp-main.207} {Knowledge prompting in pre-trained language model for natural language understanding}.
\newblock In \emph{Proceedings of the 2022 Conference on Empirical Methods in Natural Language Processing, {EMNLP} 2022, Abu Dhabi, United Arab Emirates, December 7-11, 2022}, pages 3164--3177. Association for Computational Linguistics.

\bibitem[{Wang et~al.(2024)Wang, Fang, Li, Shi, Ding, Xu, Wang, Bai, Liu, Jiayang, Chan, and Song}]{DBLP:conf/acl/0001FLS0XWBLJCS24}
Weiqi Wang, Tianqing Fang, Chunyang Li, Haochen Shi, Wenxuan Ding, Baixuan Xu, Zhaowei Wang, Jiaxin Bai, Xin Liu, Cheng Jiayang, Chunkit Chan, and Yangqiu Song. 2024.
\newblock \href {https://aclanthology.org/2024.acl-long.128} {{CANDLE:} iterative conceptualization and instantiation distillation from large language models for commonsense reasoning}.
\newblock In \emph{Proceedings of the 62nd Annual Meeting of the Association for Computational Linguistics (Volume 1: Long Papers), {ACL} 2024, Bangkok, Thailand, August 11-16, 2024}, pages 2351--2374. Association for Computational Linguistics.

\bibitem[{Wei et~al.(2016)Wei, Liu, and Li}]{DBLP:conf/acl/WeiLL16}
Zhongyu Wei, Yang Liu, and Yi~Li. 2016.
\newblock Is this post persuasive? ranking argumentative comments in online forum.
\newblock In \emph{Proceedings of the 54th Annual Meeting of the Association for Computational Linguistics, {ACL} 2016}.

\bibitem[{Xu et~al.(2023)Xu, Yang, Lin, Wang, Zhou, Zhang, and Mao}]{DBLP:journals/corr/abs-2305-14688}
Benfeng Xu, An~Yang, Junyang Lin, Quan Wang, Chang Zhou, Yongdong Zhang, and Zhendong Mao. 2023.
\newblock \href {http://arxiv.org/abs/2305.14688} {Expertprompting: Instructing large language models to be distinguished experts}.
\newblock \emph{CoRR}, abs/2305.14688.

\bibitem[{Yang et~al.(2020)Yang, Lin, Nogueira, Tsai, Wang, and Lin}]{DBLP:conf/coling/YangLNTWL20}
Jheng{-}Hong Yang, Sheng{-}Chieh Lin, Rodrigo~Frassetto Nogueira, Ming{-}Feng Tsai, Chuan{-}Ju Wang, and Jimmy Lin. 2020.
\newblock Designing templates for eliciting commonsense knowledge from pretrained sequence-to-sequence models.
\newblock In \emph{Proceedings of the 28th International Conference on Computational Linguistics, {COLING} 2020}, pages 3449--3453.

\bibitem[{Yang et~al.(2016)Yang, Yang, Dyer, He, Smola, and Hovy}]{DBLP:conf/naacl/YangYDHSH16}
Zichao Yang, Diyi Yang, Chris Dyer, Xiaodong He, Alexander~J. Smola, and Eduard~H. Hovy. 2016.
\newblock Hierarchical attention networks for document classification.
\newblock In \emph{{NAACL} {HLT} 2016, The 2016 Conference of the North American Chapter of the Association for Computational Linguistics: Human Language Technologies, San Diego California, USA, June 12-17, 2016}, pages 1480--1489.

\bibitem[{Yim et~al.(2024)Yim, Chan, Shi, Deng, Fan, Zheng, and Song}]{DBLP:journals/corr/abs-2408-02559}
Yauwai Yim, Chunkit Chan, Tianyu Shi, Zheye Deng, Wei Fan, Tianshi Zheng, and Yangqiu Song. 2024.
\newblock \href {https://doi.org/10.48550/ARXIV.2408.02559} {Evaluating and enhancing llms agent based on theory of mind in guandan: {A} multi-player cooperative game under imperfect information}.
\newblock \emph{CoRR}, abs/2408.02559.

\bibitem[{Yuan et~al.(2023)Yuan, Chen, Fu, Ge, Shah, Jankowski, Xiao, and Yang}]{DBLP:journals/corr/abs-2305-05252}
Siyu Yuan, Jiangjie Chen, Ziquan Fu, Xuyang Ge, Soham Shah, Charles~Robert Jankowski, Yanghua Xiao, and Deqing Yang. 2023.
\newblock \href {https://doi.org/10.48550/arXiv.2305.05252} {Distilling script knowledge from large language models for constrained language planning}.
\newblock \emph{CoRR}, abs/2305.05252.

\bibitem[{Zeng et~al.(2020)Zeng, Li, He, Gao, Lyu, and King}]{DBLP:conf/www/Zeng0HGLK20}
Jichuan Zeng, Jing Li, Yulan He, Cuiyun Gao, Michael~R. Lyu, and Irwin King. 2020.
\newblock \href {https://doi.org/10.1145/3366423.3380223} {What changed your mind: The roles of dynamic topics and discourse in argumentation process}.
\newblock In \emph{{WWW} '20: The Web Conference 2020, Taipei, Taiwan, April 20-24, 2020}, pages 1502--1513. {ACM} / {IW3C2}.

\end{thebibliography}
